\documentclass[9pt,journal,compsoc]{IEEEtran}
\ifCLASSOPTIONcompsoc
\usepackage[nocompress]{cite}
\else
\usepackage{cite}
\fi
\ifCLASSINFOpdf
  \usepackage[pdftex]{graphicx}
\else
 \usepackage[dvips]{graphicx}
\fi
\usepackage{amsmath}
\usepackage{algorithm,algorithmic}

\usepackage{array}

\ifCLASSOPTIONcompsoc
  \usepackage[caption=false,font=footnotesize,labelfont=sf,textfont=sf]{subfig}
\else
  \usepackage[caption=false,font=footnotesize]{subfig}
\fi
\usepackage{url}

\usepackage{hyperref}

\begin{document}
	%
	\title{Intrinsic Image Transfer for Illumination Manipulation}
	%
	%
	%
	%
	
	\author{Junqing Huang, Michael Ruzhansky, Qianying Zhang, Haihui Wang, ~\IEEEmembership{Member,~IEEE}
		\IEEEcompsocitemizethanks{\IEEEcompsocthanksitem J. Huang and M. Ruzhansky are with the Department of Mathematics: Analysis, Logic and Discrete Mathematics, Faculty of Sciences, Ghent University, Ghent 9000, Belgium (e-mails: Junqing.Huang@UGent.be; Michael.Ruzhansky@UGent.be).
		\IEEEcompsocthanksitem Q. Zhang is with the Department of Liberal Arts, Shenzhen Institute of  Information Technology, Shenzhen 518172, China (e-mail:  zhang\_qy@sz. \hfil\break jnu.edu.cn).
		\IEEEcompsocthanksitem H. Wang is with the school of Mathematical Sciences, Beihang University, Beijing 100191, China (e-mail: whhmath@buaa.edu.cn).}
		\thanks{Manuscript received XX, XXXX, 2022; revised XX, XXXX, 2022; accepted XX, XXXX, 2022. Date of publication XX, XXXX, 2022; date of current version XX, XXXX, 2022. (Corresponding author: Haihui Wang.)}
		\thanks{Recommended for acceptance by XXXX.}
		\thanks{Digital Object Identifier no. XX.XXXX/TPAMI.XXXX.XXXXXXX}
		}

	%
	%
	
	\markboth{Journal of \LaTeX\ Class Files,~Vol.~XX, No.~X, XXXX~2022}%
	{Shell \MakeLowercase{\textit{et al.}}: Bare Demo of IEEEtran.cls for Computer Society Journals}
	%

	\IEEEpubid{\makebox[\columnwidth]{\hfil 0000--0000/00/\$00.00~\copyright~2022 IEEE}%
		\hspace{\columnsep}\makebox[\columnwidth]{Published by the IEEE Computer Society\hfil}}


	\IEEEtitleabstractindextext{
		\begin{abstract}
		
		This paper presents a novel intrinsic image transfer (IIT) algorithm for image illumination manipulation, which creates a local image translation between two illumination surfaces. This model is built on an optimization-based framework composed of illumination, reflectance and content photo-realistic losses, respectively. Each loss is firstly defined on the corresponding sub-layers factorized by an intrinsic image decomposition and then reduced under the well-known spatial-varying illumination illumination-invariant reflectance prior \\knowledge. We illustrate that all losses, with the aid of an ``exemplar'' image, can be directly defined on images without the necessity of taking an intrinsic image decomposition, thereby giving a closed-form solution to image illumination manipulation. We also demonstrate its versatility and benefits to several illumination-related tasks: illumination compensation, image enhancement and tone mapping, and high dynamic range (HDR) image compression, and show their high-quality results on natural image datasets.
			
		\end{abstract}

		\begin{IEEEkeywords}
		Image illumination and reflectance, intrinsic image decomposation, intrinsic images transfer, illumination manipulation.
		\end{IEEEkeywords}
	}

\maketitle

\IEEEdisplaynontitleabstractindextext

%
\IEEEpeerreviewmaketitle

\IEEEraisesectionheading{\section{Introduction}\label{sec:introduction}}

%
%
%
%

\IEEEPARstart{I}{mage} illumination is one of the basic and important cues for both human and computer vision. Shadows reveal black while highlights make white  --- visual objects look quite different under varying illumination conditions. The study of image illumination jointing with perception problems can be dated back to Helmholtz who had interpreted that human perception of the ``intrinsic'' property of scene is regardless of the light shed on it, which is also known as ``lightness constancy''\cite{land1971lightness}. Although theories have been extensively studied over the past decades,  as Gilchrist\cite{gilchrist2006seeing} pointed out, how human determines lightness, whiteness of a scene, like vision in general, remains a mystery. Recent analysis has shown that human's perception of lighting is deeply influenced by human vision system (HVS), involving some, yet, not fully-understood mechanisms, complex interactions and feedback\cite{barrow1978recovering, land1971lightness, gilchrist2006seeing}.

The controlling of image illumination --- which is roughly referred to as image illumination manipulation in this paper, has received great attention in a variety of image processing and computer vision tasks. The representative instances include image tone mapping, image enhancement, high dynamic range (HDR) image compression, illumination compensation, and so on. Despite the different backgrounds, they in nature aim to solve a similar illumination problem --- that is, how to correct illumination conditions to bring out more visual information. It has also witnessed many efforts of different computational models to optimize illumination conditions for fine-balanced distributions to reduce the degeneration of brightness, contrast, color and saturation of images or videos. In general, they can be divided into two mainstream techniques: tone mapping operators (TMO)\cite{mantiuk2015high, morovic2001fundamentals} and Retinex-based methods\cite{barrow1978recovering,  land1971lightness}. TMO methods such as gamut mappings\cite{mantiuk2015high, morovic2001fundamentals} treat color intensities as image illumination and correct it by mapping the intensities (or hue, saturation, and so on) with specified tone-mapping operators or curves. This strategy is simple and easy to implement but may lead to visual artifacts such as noise exaggeration. In contrast, Retinex-based methods\cite{fattal2002gradient, ng2011total} tend to make an explicit intrinsic image decomposition and deal with the illumination layer individually. It is possible for them to produce plausible results with appropriate configurations, but the underlying intrinsic image decomposition is a challenging under-constrained problem, especially in the cases of lacking of sufficient prior knowledge. Notice also that both categorized methods can be interpreted in the context of Retinex theory, the core of them is how to make a faithful intrinsic image decomposition, no matter an explicit or implicit one, and then adjust the corresponding sub-layers appropriately.  Despite the great progress in the past decades, there still has considerable interest to exploit new methods to control image illumination for target purposes.

\IEEEpubidadjcol
In this paper, we investigate a general image illumination manipulation problem under the interpretation of an intrinsic image model. The main idea arises from the observation that a wide range of illumination-related tasks inherently shares a common illumination problem that can be characterized under the assumption of Retinex theory\cite{barrow1978recovering,  land1971lightness}. Differing from many traditional Retinex-based methods\cite{fattal2002gradient, ng2011total}, we provide a new paradigm to deal with per-pixel image illumination without the necessity of taking an explicit intrinsic image decomposition. Specifically, we propose a novel intrinsic image transfer (IIT) algorithm to implicitly create a local image translation between two illumination surfaces. Our IIT algorithm is built on a generalized optimization-based framework consisting of three photorealistic losses that are derived from image illumination, reflectance and content, respectively. By definition, each loss is first defined on the intrinsic layers factorized by an intrinsic image decomposition and further simplified under the well-known spatial-varying illumination and illumination-invariant reflectance prior knowledge. With a series of relaxations, all losses are directly defined on images instead of the underlying sub-layers, thereby avoiding the challenging intrinsic image decomposition. As explained in Section \ref{sec:methodology} and \ref{experimental_results}, the proposed IIT method has a powerful ability in preserving local consistent structures while suppressing the potential artifacts such as halos around the salient edges and textural distortions, which provides an easy-configurated illumination manipulation tool for many practical applications. The major contributions of this paper are summarized as follows:  

\begin{itemize}
	\setlength{\itemsep}{0pt}
	\setlength{\parsep}{0pt}
	\setlength{\parskip}{1pt}
	
	\item A new generalized minimization framework--- intrinsic image transfer (IIT), is concisely designed for image illumination manipulation, in which three photorealistic losses (illumination, reflectance and content) are defined to characterize the underlying ``intrinsic'' layers. 
	
	\item A simple filtering operator is introduced to mimic the spatial-smoothing property of image illumination, and a locally linear embedding (LLE) algorithm\cite{roweis2000nonlinear} is also formulated to identify the illumination-invariant reflectance. The refinements provide an alternative way to optimize image illumination without the necessity of taking an explicit intrinsic image decomposition.
	
	\item A closed-form solution to controlling image illumination is implemented with the aid of a so-called exemplar image. The performance is verified both qualitatively and quantitatively on several	illumination-related tasks, where our IIT algorithm performs favorable results on natural images against state-of-the-art methods.
	
\end{itemize}

We further remark that our IIT method takes a small step toward clarifying the internal relationship between ``intrinsic'' layers and a wide range of image illumination-related tasks. It is in nature built on intrinsic images, but it does not strive to explore an explicit intrinsic image decomposition. Instead, we implicitly characterize and regularize the underlying ``intrinsic'' image layers based on the well-known spatially-smoothing illumination and illumination-invariant reflectance assumption. The experimental results also demonstrate its versatility and many potential benefits to illumination compensation, image enhancement, HDR image compression, and so on.

The rest paper is organized as follows. The related work is presented in Section \ref{sec:related_work}. The proposed method is presented in Section \ref{sec:methodology}, in which a generalized intrinsic image model and its connections to image illumination manipulation are discussed in Section \ref{subsec:intrinsic_model}  and \ref{subsec:multiple_observations}, respectively; and the model together with three photorealistic losses is systemically explored in Section \ref{subsec:iit_model}. In Section \ref{experimental_results}, a series of experimental results are presented with the quantitative evaluation on different image datasets. In Section \ref{conclusion}, we draw our concluding remarks and future directions.

\section{Related Work}\label{sec:related_work}

Previous work aiming to find a fine-balanced illumination distribution covers a wide range of techniques. We here mainly review the related methods based on the underlying ``intrinsic'' images and illustrate the connections and potential benefits to image illumination manipulation.

Retinex theory is a famous computational model proposed by Land and McCann\cite{land1971lightness} to interpret the ``lightness constancy'' phenomenon, which is also an important theoretical guide for image illumination manipulation. In Retinex theory, it is usually assumed a Mondrian-like image that can be formulated by the point-wise multiplication of two components (illumination and reflection layers). The illumination layer measures the amount of incident light illuminated on an object or scene; while the reflectance layer depicts the reliable visual information independent with the varying illumination. Land\cite{land1977retinex} also derived a simple algorithm that classifies the strong gradients of an image into reflectance layer and all the other variations into illumination layer. In\cite{horn1974determining}, Horn interpreted that a complete image decomposition of them can be achieved by solving a Poisson equation mathematically. Due to its effectiveness, Retinex-based models have been widely used as basic tools for a variety of illumination-related tasks such as image tone mapping and enhancement, image illumination compensation, HDR image compression, and so on. For instance, Jobson\cite{land1983recent} explored the properties of the center/surround function and proposed a so-called center/surround Retinex for HDR image compression. The work was subsequently extended to the classic multiscale Retinex\cite{jobson1997multiscale} for color, contrast and brightness correction. Fattal et al.\cite{fattal2002gradient} proposed an HDR image compression method based on Retinex theory in the gradient domain of images. Besides, Retinex theory has also stimulated many other computational models, including the variational-based image decomposition\cite{kimmel2003variational}, PDE-based image smoothing\cite{morel2010pde}, to clarify the relationship of two components, as well as the benefits to these aforementioned illumination-related tasks. Besides, Elad\cite{elad2005retinex} justified the relationship between a regularized Retinex model and bilateral filter\cite{tomasi1998bilateral} theoretically by characterizing the smoothing illumination, which also stimulates the use of edge-ware image filters for image decomposition and illumination manipulation. The interested reader is referred to the  surveys\cite{farbman2008edge, tomasi1998bilateral, milanfar2012tour} for more details.

Simultaneously, intrinsic image decomposition has also drawn great attention since the terminology ``intrinsic images'' was introduced by Barrow and Tenenbaum\cite{barrow1978recovering}. The methods essentially share very similar assumptions as Retinex-based methods, while they strive to decompose an image into different ``intrinsic'' components explicitly. In general, it is a highly under-constrained problem to recover the multiple intrinsic layers from single or multiple images without any further assumptions. In order to obtain a faithful image decomposition, Rother, et al.\cite{rother2011recovering}, for example, gave an image decomposition method by encoding the sparsity of the reflectance with a Gaussian mixture model.  Bell et al.\cite{bell2014intrinsic} presented a dense conditional random field algorithm for intrinsic image decomposition on a large-scale indoor dataset. Recently, more complex computational models were proposed for real-world image decomposition. In\cite{lombardi2016radiometric}, depth cues with RGB-D cameras were used to infer the intrinsic images. Jeon et al.\cite{jeon2014intrinsic} introduced an image decomposition model based on locally linear embedding algorithm\cite{roweis2000nonlinear}, which is related to our model but mainly strives to explicitly decompose an image into shading, reflectance and texture layers. Barron and Mailk\cite{barron2015shape} also formulated a similar one called ``shape, albedo, and illumination from shading (SAIFS)'' to recover each intrinsic layer from an image. Instead, Guo et al.\cite{guo2017lime} proposed a model for low-light image enhancement that takes only the illumination layer into account under dark-channel prior. When it comes to image illumination manipulation, intrinsic image decomposition is always used as a building block integrated with a layer-remapping operator and image construction for post-processing. Nevertheless, the post-processing steps are out of the constraints of intrinsic image decomposition, which may introduce strong artifacts even with a high-quality image decomposition.

\begin{figure*}[t]
	\begin{center}
		\begin{tabular}{c}
			\begin{minipage}[t]{0.50\textwidth}
				\subfloat[Input]
				{\includegraphics[width=0.2\textwidth]{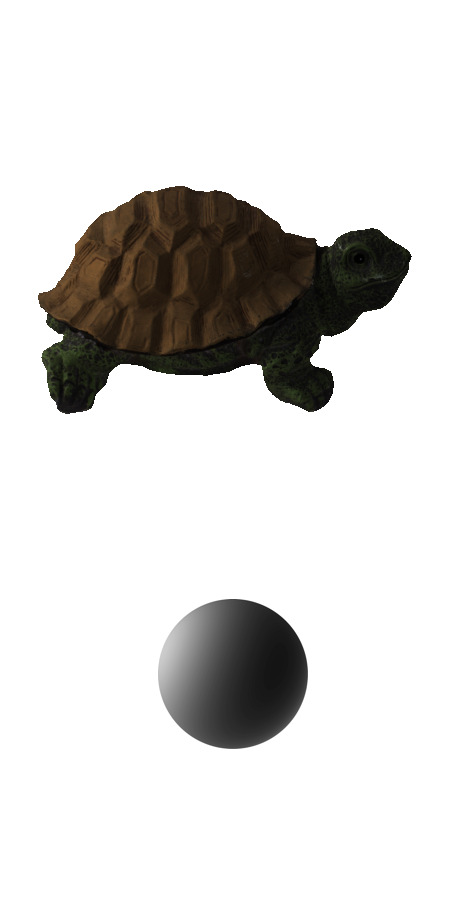}}
				\subfloat[ Illumination and reflectance mapping]
				{\includegraphics[width=0.8\textwidth]{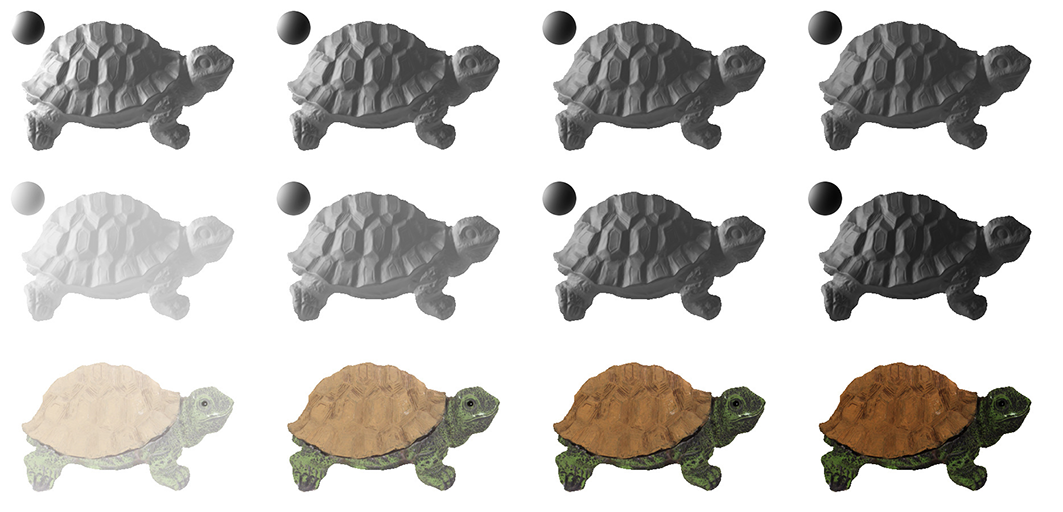}}
			\end{minipage}
		\end{tabular}
		\begin{tabular}{|c}
			\begin{minipage}[t]{0.42\textwidth}
				\subfloat[Input]
				{\includegraphics[width=0.2\textwidth]{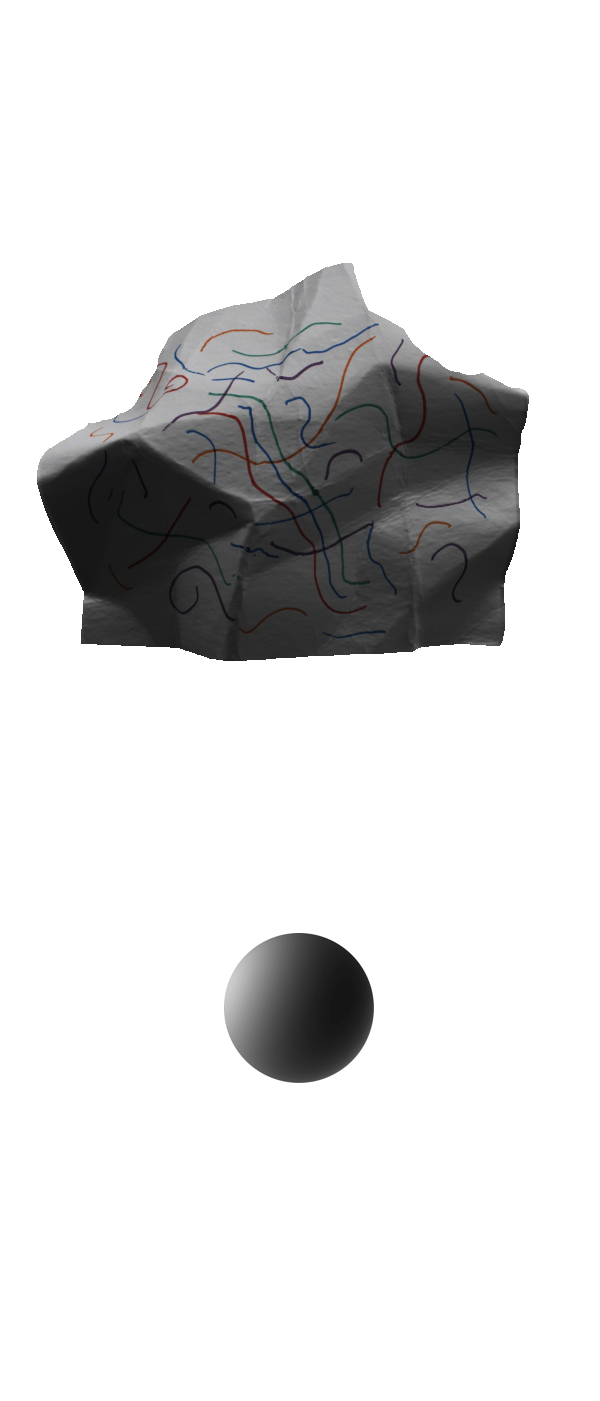}}
				\subfloat[ Illumination and reflectance mapping]
				{\includegraphics[width=0.8\textwidth]{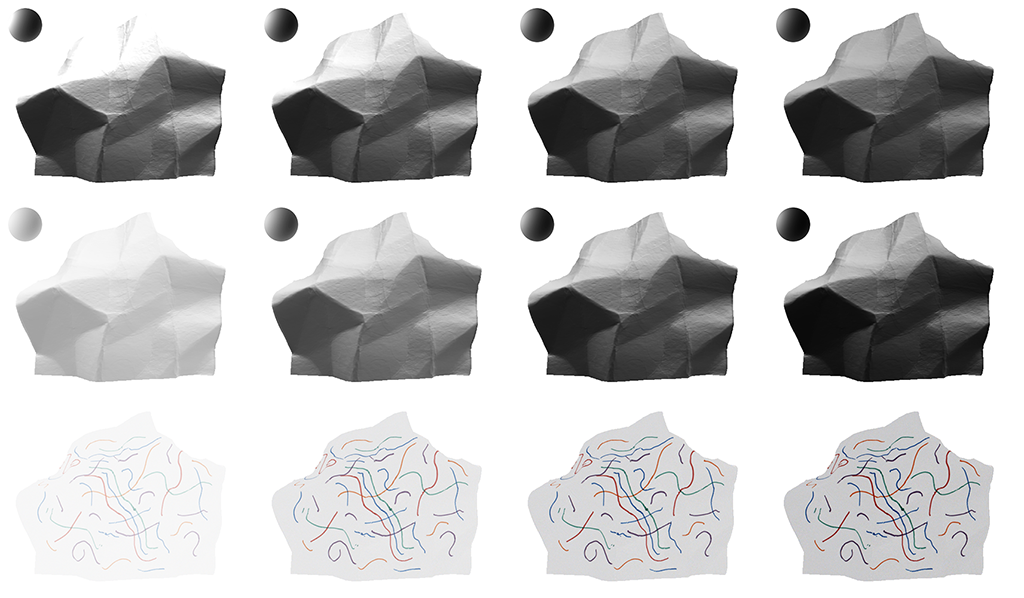}}
			\end{minipage}
		\end{tabular}
	\end{center}
	\caption{ Illumination and reflectance mapping in the logarithm domain.  Top: local illumination mapping (${a=[0.3, 0.1, -0.1, -0.3]}$), middle: global illumination mapping (${b=[0.6, 0.9, 1.1, 1.4]}$), bottom: global reflectance mapping (${c=[0.5, 0.8, 1.2, 1.5]}$).
	}\label{fig:fig1}
\end{figure*}

Besides, recent advances in learning-based methods achieve a milestone for the overwhelming results in many challenging benchmarks\cite{bell2014intrinsic, Cordts2016Cityscapes, geiger2013vision, ignatov2017dslr, krizhevsky2012imagenet}. Strongly benefiting from the deep convolutional neural networks (CNN)\cite{goodfellow2014generative, krizhevsky2012imagenet}, more delicate network architectures are springing out. It has also witnessed the use of deep learning approaches\cite{cheng2018intrinsic, gardner2017learning, hold2017deep, li2018learning, meka2018lime, wei2018deep} to explore and infer the intrinsic layers from single or multiple images or directly recover image illumination for better visual perception or interpretation. For example, Narihira et al.\cite{narihira2015direct} first introduced deep learning to learn albedo from MPI Sintel dataset\cite{butler2012naturalistic}. Chen et al.\cite{chen2018learning} recently released an end-to-end training model with a fully convolutional network for the low-light images by using raw images for training.  Andrey et al.\cite{ignatov2017wespe} presented an end-to-end weakly supervised photo enhancer (WESPE) to enhance an image automatically. Yet, one of the bottlenecks of these deep learning-based methods is the necessity of a large scale of high-quality matched training pairs for supervision. More recently, unsupervised image-to-image translation\cite{guo2020zero} bring new hope and replace parts of the traditional pipelines for easy and flexible real-world image tone mapping and enhancement\cite{chen2018deep, janner2017self, jiang2019enlightengan,  zhu2017unpaired, zhang2021star}. The reader is referred to these existing deep learning-based algorithms for more details.

\section{Methodology}\label{sec:methodology}

In this section, we briefly introduce a well-known intrinsic image model and show how to derive the proposed IIT model and reduce the photorealistic losses under the intrinsic images mathematically.

\subsection{Intrinsic Images Model}\label{subsec:intrinsic_model}
In many existing Retinex-based model\cite{barrow1978recovering, rother2011recovering}, an image $\mathcal{I}$ is assumed to be factorized into illumination $\mathcal{L}$ and reflectance $\mathcal{R}$,
\begin{equation}
	\label{eq:eq1}
	\begin{aligned}
		\mathcal{I=L\odot R,}
	\end{aligned}
\end{equation}
where ${\odot}$ is a point-wise multiplication operator, $\mathcal{L}$ represents the light-dependent properties such as shading, shadows or specular highlights of images, and $\mathcal{R}$ represents the material-dependent properties, known as the reflectance of a scene. $\mathcal{L}$ and $\mathcal{R}$ take rather different roles in controlling the image color, contrast, brightness and so on. Such an image decomposition has formed a basis for many intrinsic image decomposition methods\cite{bell2014intrinsic, horn1974determining, land1977retinex, rother2011recovering}.
 
It is clear that Eq. \ref{eq:eq1} is a highly under-constraint problem if no further assumption is imposed on illumination $\mathcal{L}$ or reflectance $\mathcal{R}$. Previous work mainly focuses on exploring sensible priors --- for example, the well-known spatial-varying illumination and illumination-invariant reflectance --- that is, illumination $\mathcal{L}$ is assumed to be spatial-smoothing and determines the brightness of a scene, while reflectance $\mathcal{R}$  possesses illumination-invariant property and is expected to be piece-wise constant under varying lighting conditions. Such a prior may be further interpreted as sparse priors\cite{guo2017lime, shen2011intrinsic}, low-rank\cite{huang2018multispectral,zheng2015illumination}, non-local cues\cite{zhao2012closed, zosso2015non} and so on. Despite many efforts, it is still challenging to make an explicit intrinsic image decomposition, especially for natural images under complex light, material, color and geometry conditions.

\subsection{Multiple Observations}\label{subsec:multiple_observations}

We claim that a wide range of image illumination-related tasks, for example,  image illumination compensation, tone mapping, enhancement and HDR image compression, can be understood in the context of Eq. \ref{eq:eq1}. For example, it is always preferable in many tone mapping methods\cite{mantiuk2015high, morovic2001fundamentals} to treat color intensities as illumination ${\mathcal{L}}$ and correct illumination by applying tonal adjustments such as linear contrast stretching and gamma correction on image intensities. In Retinex-based methods\cite{barrow1978recovering,  land1971lightness}, a more complex procedure: intrinsic image decomposition, layer-remapping and image reconstruction may be applied for more plausible results. Despite the different backgrounds and generalizations, we can see that they essentially share a common illumination problem --- that is, how to decouple image illumination from images or make an explicit intrinsic image decomposition, and then adjust each layer appropriately to achieve the target purpose.

Suppose a high-quality intrinsic image decomposition is available, we depict the role of layer-remapping operators in aforementioned illumination-related tasks. As suggested in\cite{rother2011recovering}, we consider a simple linear mapping model ${\hat{I} \!=\! a \!+\! bL \!+\! cR}$ in the logarithmic domain, where ${\emph{I}}$, ${\emph{L}}$ and ${\emph{R}}$ are the logarithmic counterparts: ${\mathcal{I}}$, ${\mathcal{L}}$ and ${\mathcal{R}}$, respectively. It is clear in Fig.  \ref{fig:fig1} that $a$ mainly controls the local brightness and contrast of illumination ${\emph{L}}$ and $b$ plays a similar role but acts globally, while $c$ affects the global brightness and contrast of the reflectance ${\emph{R}}$\footnote{It is usually to keep $c=1$ invariant unless the reflectance layer ${\emph{R}}$ needs to be adjusted in some situations because of the illumination-independent nature.}.  Obviously, the visual results can be greatly affected by the parameters $a, b$ and $c$. In practice, such layer-remapping operators may be implemented in a more complex form, for example, using a spatial-varying or adaptive mapping function for high-quality image illumination manipulation results. No matter in what form, such a strategy has limitations due to the facts: (a) intrinsic image decomposition is a highly under-constrained problem, and the estimation of each sub-layer highly relies on the prior knowledge; (b) it is not very easy to determine an appropriate layer-remapped operator for the sub-layers even with a high-quality image decomposition, and (c) the artifacts introduced by layer-remapping and image construction are not easy to control for out of the image decomposition. 

As illustrated hereafter, the drawbacks can be significantly alleviated by integrating the intrinsic image decomposition, layer-remapping and image reconstruction into a generalized optimization-based framework. Such a strategy enables us to control image illumination in an implicit way --- that is, to regularize each sub-layer without the necessity of taking an explicit intrinsic image decomposition. The advantages mainly arise from the well-known spatial-varying illumination and illumination-invariant reflectance prior knowledge. We briefly list them as follows:

\begin{itemize}
	\setlength{\itemsep}{0pt}
	\setlength{\parsep}{0pt}
	\setlength{\parskip}{0pt}
	
	\item Spatial-smoothing illumination --- that is, illumination $L$ has the spatial-smoothing property, while the majority of dramatic variations such as strong or salient edges, textures and structures are attributed to reflectance ${R}$. 
	
	\item Illumination-invariant reflectance --- that is, reflectance $R$ tends to be invariant under varying illumination conditions. It is straightforward to claim that the local geometric structures formed by the dramatic variations in $R$ have the same illumination-invariant property. 
	
	\item The visibility of an image is primarily determined by illumination ${L}$, the intensities of which may be, locally or globally, compensated or corrected for its discrepancy to the target one, especially under varying illumination conditions. Human vision system is not so sensitive to the absolute change of illumination as that of reflectance due to the ``color-consistency'' phenomenon.  
	
\end{itemize}

We hereafter illustrate the benefits of three assumptions for simplifying the process of image illumination manipulation. As interpreted in Section \ref{subsec:iit_model}, the first assumption motivates us to use smoothing filters to approximate image illumination, because the abundant dramatic varying features are divided into reflectance. The second one implies that the illumination-invariant property of image reflectance can be constrained by regularizing the varying features equivalently. The last one indicates that image illumination should be approximated to a balanced one under the degradation illumination situations. More details are discussed later in this paper.

\subsection{Illumination Manipulation Model}\label{subsec:iit_model}
We now elaborate how to regularize the intrinsic images given by Eq. (1). The motivation here is to design a simple, yet, powerful model for image illumination manipulation without the necessary of taking an explicit image decomposition.

Let ${\boldsymbol{s}=\{{s}_{i}\}_{i=1}^N}$ and ${\boldsymbol{o}=\{{o}_{i}\}_{i=1}^N}$ be the input and output images with ${s}_{i}$ and ${o}_{i}$ denoted as pixel intensities, respectively, the output image ${\boldsymbol{o}}$ can be expressed as an optimal solution of the following minimization problem,
\begin{equation}
	\label{eq:eq2}
	\begin{aligned}
		\mathop{\min}_{\boldsymbol{o}} E(\boldsymbol{o}) = {\alpha} E^l(\boldsymbol{o}) + {\beta} E^r(\boldsymbol{o}) + {\gamma}E^c(\boldsymbol{o}),
	\end{aligned}
\end{equation}
where photorealistic loss $E$ includes three terms ${E^l,E^r}$ and ${E^c}$ which are defined on illumination, reflectance and content respectively. ${{\alpha}, {\beta}, {\gamma}}$ control the balance of three terms.

\textbf{Illumination loss:} \quad In many practical illumination-related tasks, it may suffer from great image illumination degradation due to the varying illumination condition. In this case, it is always desirable to restore the degraded illumination into a fine-balanced one. Suppose that an extra image ${\boldsymbol{c}=\{{c}_{i}\}_{i=1}^N}$ has the ideal latent illumination ${\boldsymbol{c}^l}$, it is appropriate to impose the output illumination ${\boldsymbol{o}^l}$ to be close to the latent illumination ${\boldsymbol{c}^l}$, leading to the illumination loss ${E^l(\boldsymbol{o})}$,
\begin{equation}
	\label{eq:eq3}
	\begin{aligned}
		E^l(\boldsymbol{o})= {\sum_{i} {({o}_{i}^{l}-{c}_{i}^{l})}^2},
	\end{aligned}
\end{equation}
where ${{o}_{i}^{l}}$  and ${{c}_{i}^{l}}$  are the ${i}$-th pixel illumination of output and latent images, respectively; and the up-index $l$ represents that Eq. \ref{eq:eq3} is defined on the illumination layers. An ideal image $c$ is obviously not available in advance, but, as we illustrate hereafter, it is possible to be approximated with the aid of a so-called ``exemplar'' image in terms of its role in guiding the image illumination. 

Notice that the definition of Eq. \ref{eq:eq3} needs a complex intrinsic image decomposition. Recalling the spatial-smoothing property of illumination $L$, it is possible to apply a filter on images and treat the smoothing results as the illumination layers. Considering a simple Gaussian-like smoothing kernel ${\mathcal{K}}$, the illumination ${{o}_{i}^{l}}$ can be represented as, 
\begin{equation}
	\label{eq:eq4}
	\begin{aligned}
		{o}_{i}^{l}&={\sum_{j\in{\mathcal{N}_{i}}} \mathcal{K}_{i,j} {o}_{j}}, \quad {\sum_{j\in{\mathcal{N}_{i}}} \mathcal{K}_{i,j}=1},\\
		&\mathcal{K}_{i,j}\propto{\text{exp}\left(-\frac{1}{\delta_{\boldsymbol{f}}^2}{\left\Vert \boldsymbol{f}_i-\boldsymbol{f}_j \right\Vert }_2^2\right)},
	\end{aligned}
\end{equation}
where ${\mathcal{N}_{i}}$ denotes a neighbor set of pixel ${i}$, ${\mathcal{K}_{i,j}}$ is the Gaussian kernel weight between the pixel ${i}$ and ${j}$; and ${\boldsymbol{f}}$ is a feature vector with standard deviation ${\delta_{\boldsymbol{f}}}$. ${\boldsymbol{f}}$ can be the pixel's position, intensity, chromaticity or abstract features. We suggest that ${E^l(\boldsymbol{o})}$ can be defined as,
\begin{equation}
	\label{eq:eq5}
	\begin{aligned}
		E^l(\boldsymbol{o})= {\sum_{i}\sum_{j\in{\mathcal{N}_{i}}}(\mathcal{K}_{i,j}^o {o}_{j}-\mathcal{K}_{i,j}^c {c}_{j})^2},
	\end{aligned}
\end{equation}
where ${\mathcal{K}_{i,j}^o}$ and ${\mathcal{K}_{i,j}^c}$ are corresponding filter kernels. 
In this case, Eq. \ref{eq:eq5} plays an identical role as Eq. \ref{eq:eq3}  in measuring the illumination loss under the spatial-smoothing property. The benefit of introducing the filter kernel ${\mathcal{K}}$ is to simplify the illumination loss without taking an explicit image decomposition. 

\textbf{Reflectance loss:} \quad Recall the illumination-invariant property of reflectance layers, it is highly reasonable to expect the output reflectance ${\boldsymbol{o}^r}$ to be identical to ${\boldsymbol{s}^r}$, that of original image. In such a sense, we define ${E^r(\boldsymbol{o})}$ as,
\begin{equation}
	\label{eq:eq6}
	\begin{aligned}
		E^r(\boldsymbol{o})=  {\sum_{i} {({o}_{i}^{r}-{s}_{i}^{r})}^2},
	\end{aligned}
\end{equation}
where ${{s}_{i}^{r}}$ and ${{o}_{i}^{r}}$ are the ${i}$-th pixels values of the corresponding reflectance layers. Again, Eq. \ref{eq:eq6} requires an ill-posed intrinsic image decomposition. Notice that the reflectance layer $R$ is assumed to have abundant features such as salient edges, lines and textures and have the same illumination-invariant property, which motivates us to control $R$ by regularizing these features. Mathematically, we use a local linear model to encode these features, that is,  each point of the reflectance layer can be expressed as a weighted sum of its neighbors,

\begin{equation}
	\label{eq:eq7}
	\begin{aligned}
		s^r_i={\sum_{j \in \Omega_{i}} {\omega_{i,j}^{s^r} {s^r_j}}},
	\end{aligned}
\end{equation}
where ${\Omega_{i}}$ is a neighbor of pixel ${i}$ and the weight ${\omega_{i,j}^{s^r}}$ satisfies ${\sum_{j \in \Omega_{i}} {\omega_{i,j}^{s^r}}=1}$. 
It is plausible that if the reflectance ${\boldsymbol{s}^r}$ keeps invariant, the weight ${\omega_{i,j}^{s^r}}$ is invariant as well. By analogy with the output ${\boldsymbol{o}^r}$, the photorealistic reflectance loss can be reformulated into the following constraints equivalently,
\begin{equation}
	\label{eq:eq8}
	\begin{aligned}
		\begin{cases}
			&{o_i^r}= {\sum_{j \in \Omega_{i}}{\omega_{i,j}^{o^r}}{o_j^r}}, \quad \sum_{j \in \Omega_{i}}{\omega_{i,j}^{o^r}}=1,\\
			&{{s_i^r}}= {\sum_{j \in \Omega_{i}}{\omega_{i,j}^{s^r}}{s_j^r}}, \quad \sum_{j \in \Omega_{i}} {\omega_{i,j}^{s^r}}=1,\\
			&{\omega_{i,j}^{o^r}} = {\omega_{i,j}^{s^r}}.
		\end{cases}
	\end{aligned}
\end{equation}
where ${{\omega_{i,j}^{o^r}}}$ and ${{\omega_{i,j}^{s^r}}}$ represent the weights, and the up-index ${s^r}$ and ${o^r}$ denote that the local linear ``encoding" operators are acted on the reflectance layers. In the Eq. \ref{eq:eq8}, we force ${{\omega_{i,j}^{o^r}}={\omega_{i,j}^{s^r}}}$  to give a structural-consistency constraint for the reflectance layers. 
\begin{figure}[t]
	\begin{center}
		\begin{minipage}{0.48\textwidth}
			\centering
			\includegraphics[width=0.24\textwidth]{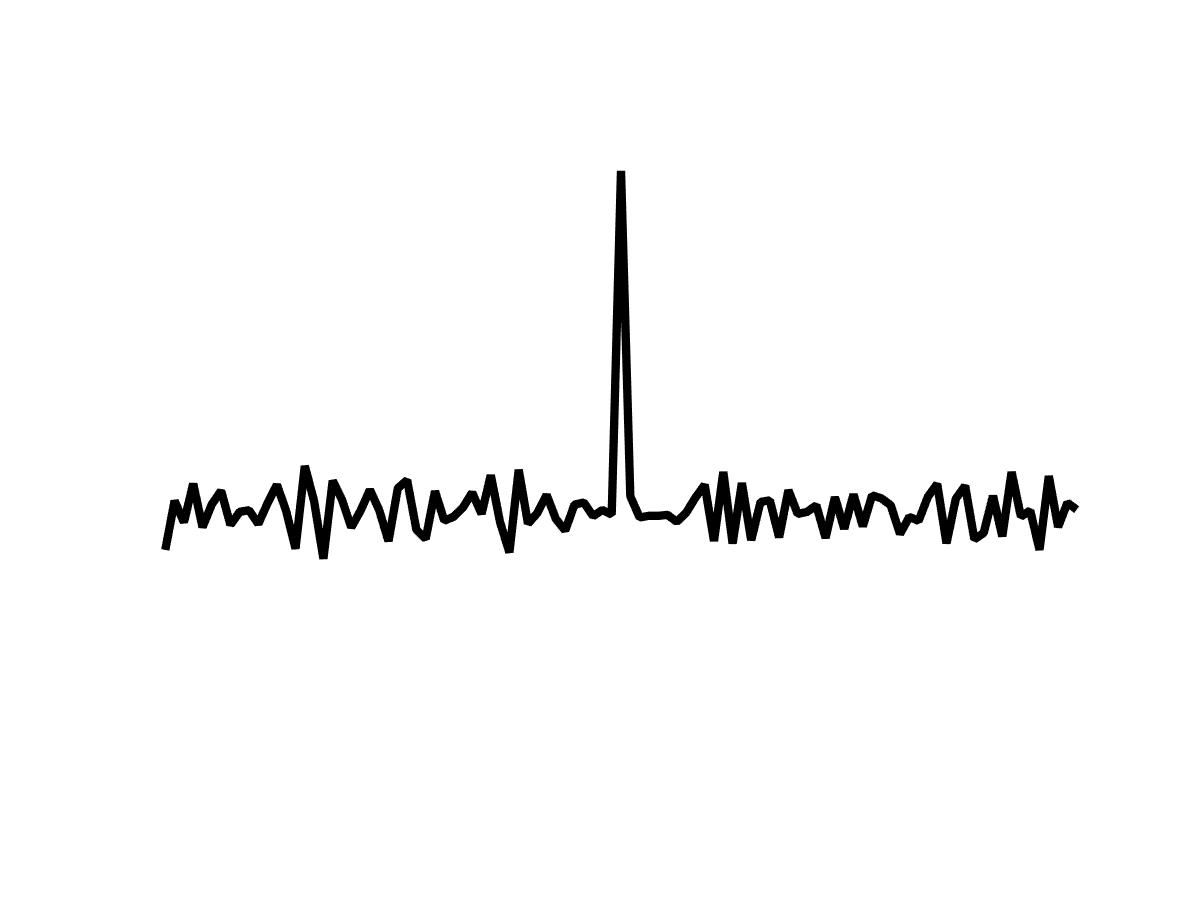}
			\includegraphics[width=0.24\textwidth]{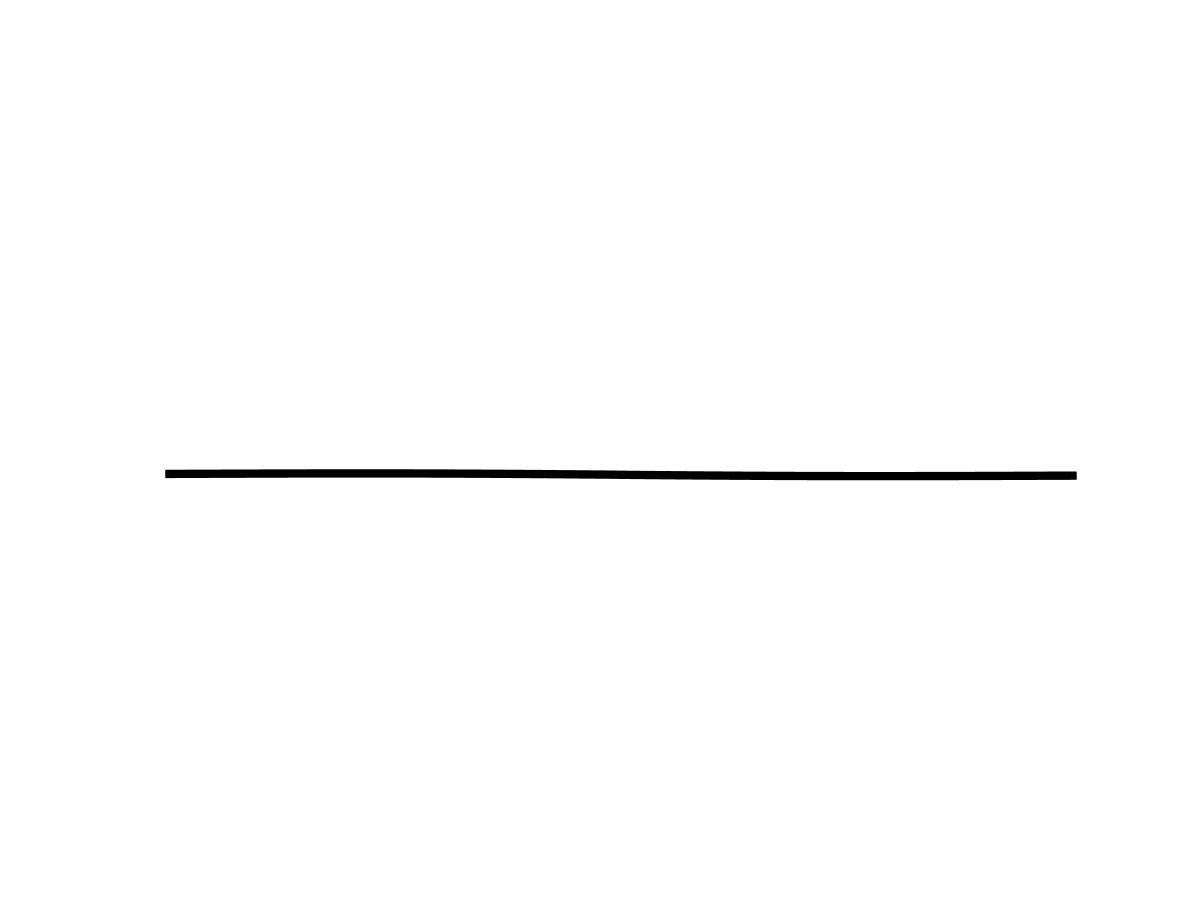}
			\fbox{
					\includegraphics[width=0.2\textwidth]{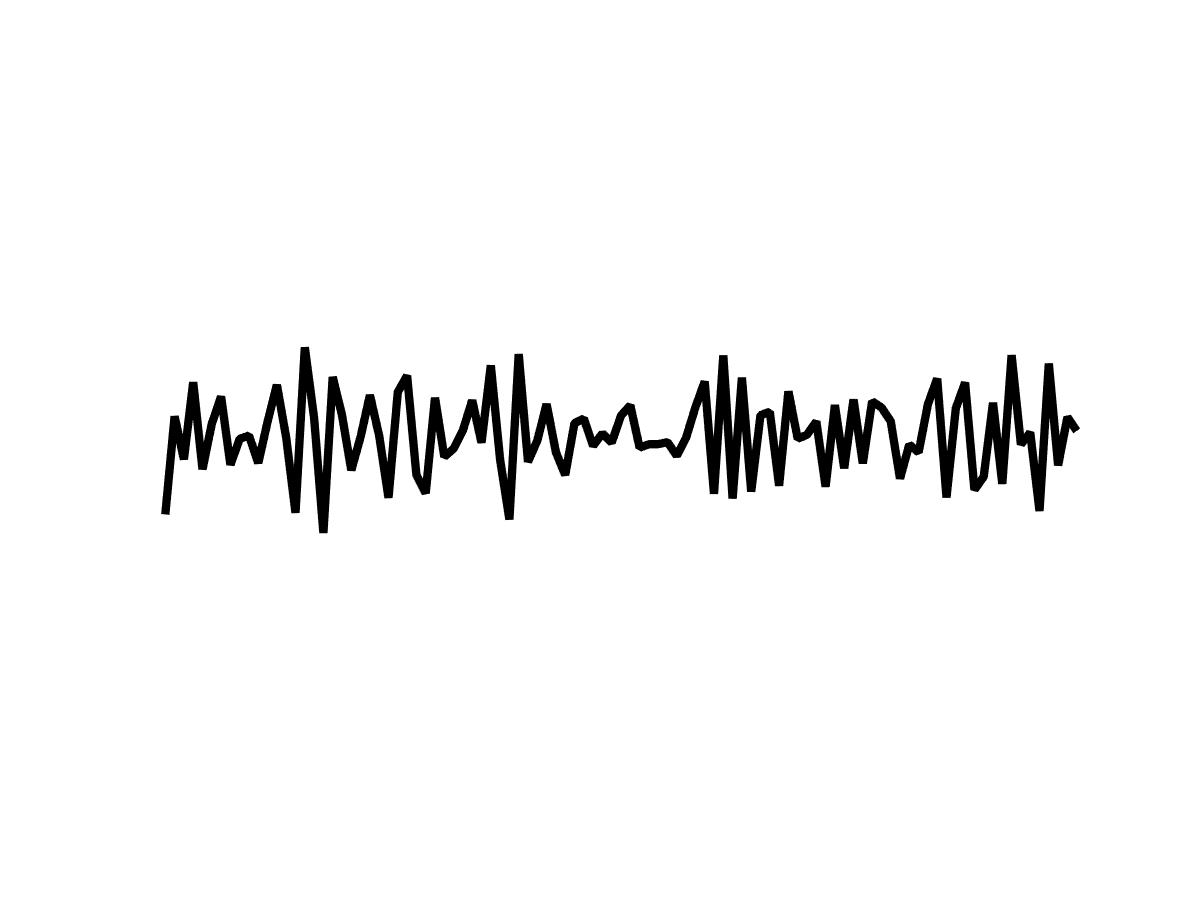}
					\includegraphics[width=0.2\textwidth]{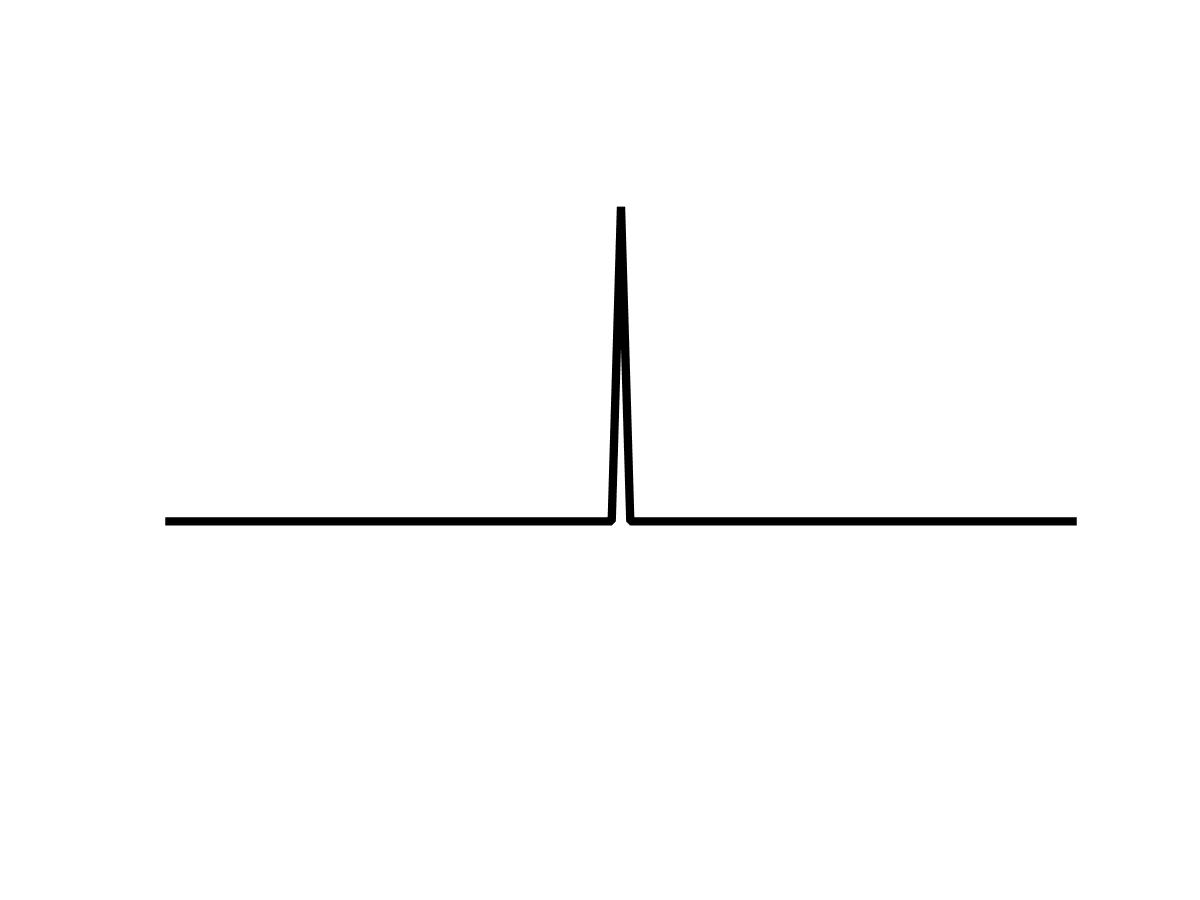}
			}
			\subfloat[Input ${I}$]
			{\includegraphics[width=0.24\textwidth]{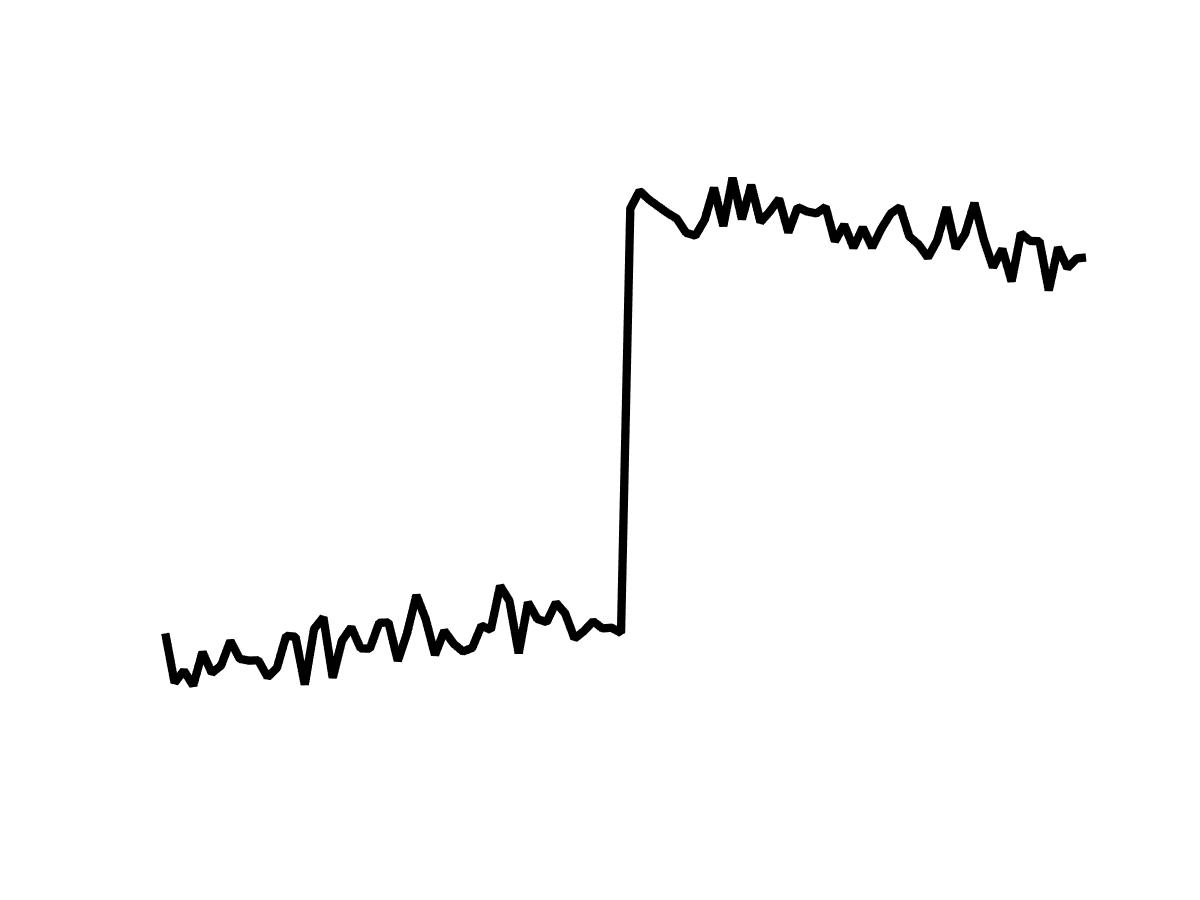}}
			\subfloat[Layer ${L}$]
			{\includegraphics[width=0.24\textwidth]{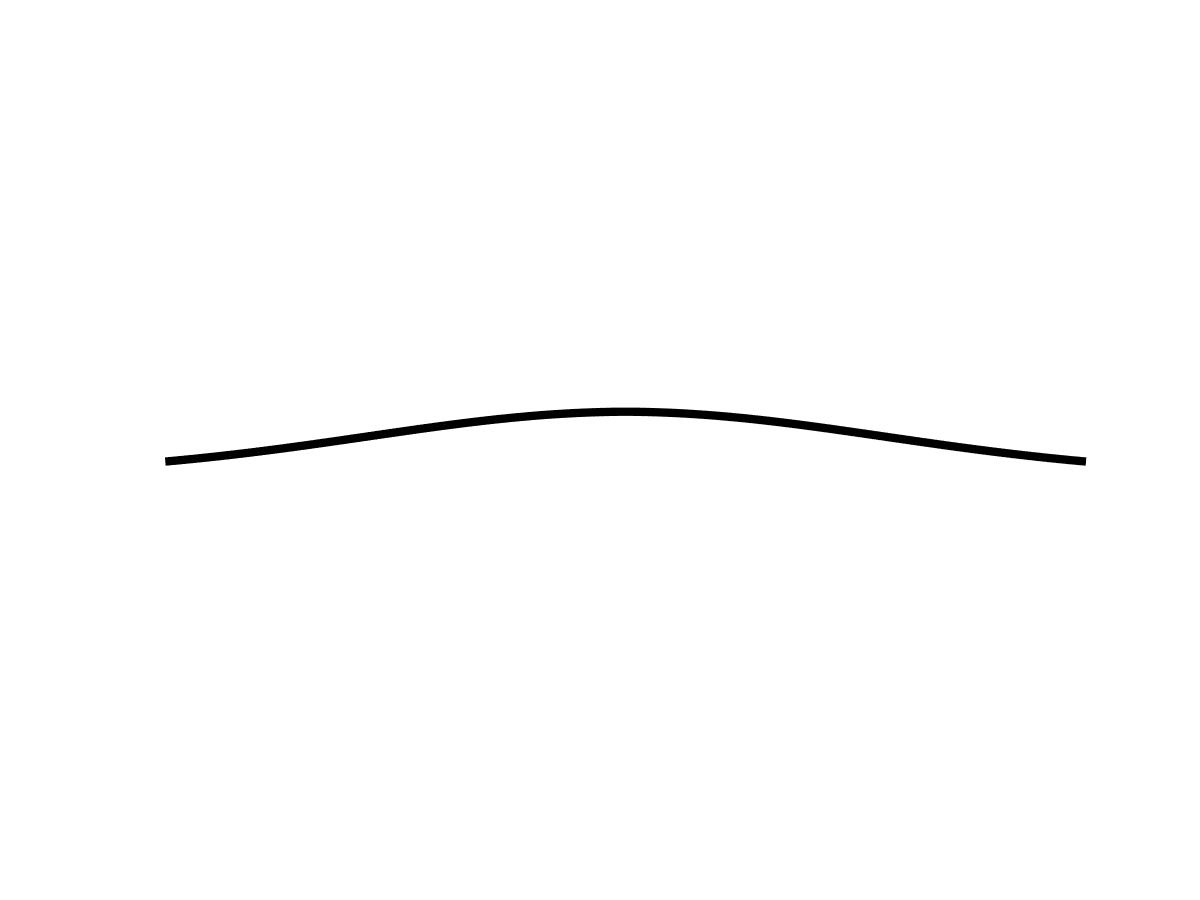}}
			\subfloat[Layer ${R}$]
			{
				\fbox{
					\includegraphics[width=0.2\textwidth]{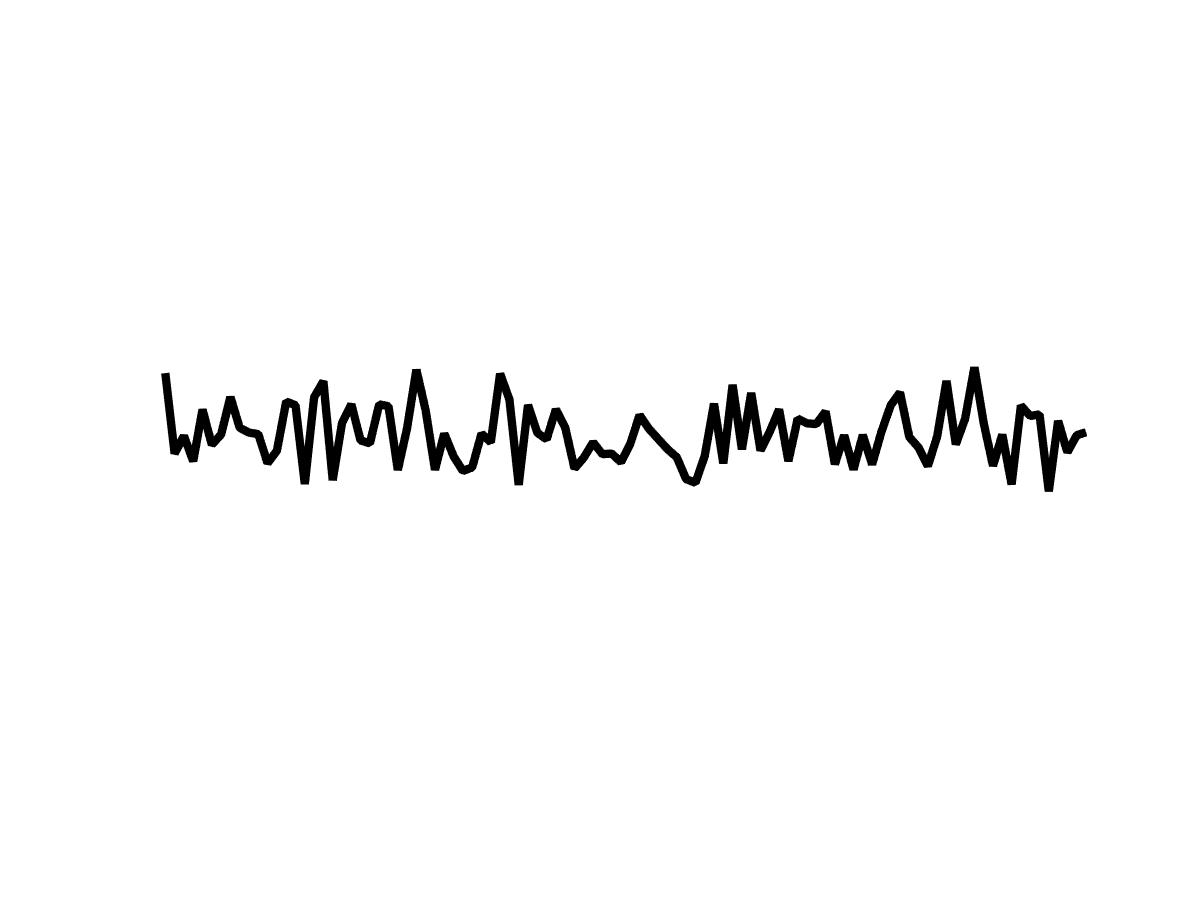}
					\includegraphics[width=0.2\textwidth]{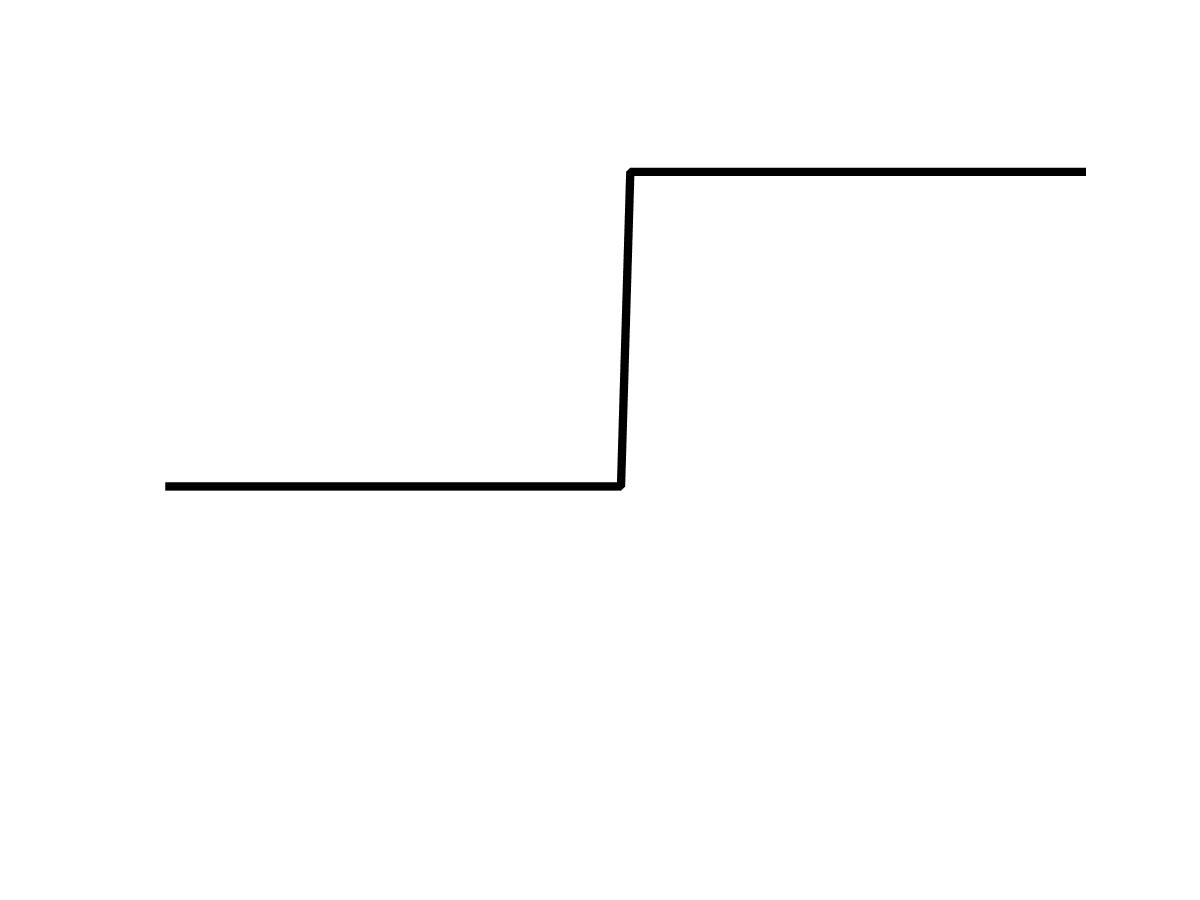}
				}
			}
		\end{minipage}
	\end{center}
	\caption{The ${1}$\text{-D} signal ${I}$ is decomposed into three basic components: a spatial-varying ${L}$,  a detail texture layer and a salient edge.  We attribute the texture and edge into the reflectance ${R}$ in our intrinsic image model.}
	\label{fig:fig2}
\end{figure} %
Recall the spatial-varying illumination, we have ${{{s}^{l}_{k}} \approx \bar{{s}^{l}_{k}}}$ with mean value ${\bar{{s}^{l}_{k}}}$ in local patch ${k}$. Besides, if two patches ${i}$ and ${j}$  are close to each other, we also have ${\bar{{s}^{l}_{i}} \approx \bar{{s}^{l}_{j}}}$. As shown in Fig.  \ref{fig:fig2},  the two claims are valid in both flat regions (Top) and strong edges (Bottom). Substituting them into Eq. \ref{eq:eq7}, we can further simplify the ``encoding'' model into,
\begin{equation}
	\label{eq:eq9}
	\begin{aligned}
		{s}_{i} &={({s}^{r}_{i}+{s}^{l}_{i})} \approx {({s}^{r}_{i}+\bar{{s}^{l}_{i}})} = {({\sum\nolimits_{j \in \Omega_{i}}{\omega_{i,j}^{s^r}}{{s}^{r}_j}}+\bar{{s}^{l}_{i}})} \\
		&= {{\sum\nolimits_{j \in \Omega_{i}}{\omega_{i,j}^{s^r}}({{s}^{r}_j}}+\bar{{s}^{l}_{i}})} \approx {{\sum\nolimits_{j \in \Omega_{i}}{\omega_{i,j}^{s^r}}({{s}^{r}_j}}+\bar{{s}^{l}_{j}})} \\
		&\approx {{\sum\nolimits_{j \in \Omega_{i}}{\omega_{i,j}^{s^r}}({{s}^{r}_j}}+{{s}^{l}_{j}})}={\sum\nolimits_{j \in \Omega_{i}}{\omega_{i,j}^{s^r}}{{s}_j}}.
	\end{aligned}
\end{equation}

As we can see, Eq. (9) and (7) have the same form and ${\omega_{i,j}^{s^r}}$ keeps invariant when adding a constant illumination ${\bar{{s}^{l}_{i}}}$ back. Due to the translation-invariant property of local linear model, the reflectance layer can be directly regularized without the necessity of decoupling the reflectance from an image. Putting Eq. (7) and (9) into Eq. (8), we have reflectance loss ${E^r(\boldsymbol{o})}$,
\begin{equation}
	\label{eq:eq10}
	\begin{aligned}
		&E^r(\boldsymbol{o})={\sum_{i} {(o_i-{\sum_{j \in \Omega_{i}}{\omega_{i,j}^{o}}{o_j}})}^2}\\
		\text{s.t.}&\quad {\omega_{i,j}^{o}}= {\omega_{i,j}^{s}}, \quad {s_i}= {\sum_{j \in \Omega_{i}}{\omega_{i,j}^{s}{s_j}}},
	\end{aligned}
\end{equation}
where ${\omega_{i,j}^{s}}={\omega_{i,j}^{s^r}}$ and ${\omega_{i,j}^{o}}={\omega_{i,j}^{o^r}}$ for local ``encoding" weights. The relationship between Eq. \ref{eq:eq8} and Eq. \ref{eq:eq10} is clear, as the first constraint in Eq. \ref{eq:eq8} is chosen as the objective function and the others act as constraints. As we can see, Eq. \ref{eq:eq10} contributes an identical regularizing role as  Eq. \ref{eq:eq6}. The simplification roots in the translation-invariant property of locally linear embedding (LLE)\cite{roweis2000nonlinear}, which helps to faithfully penalty non-consistent structures despite the local illumination deviation between the input and exemplar images. Moreover, this reduction enables us to translate the reflectance layer from the source illumination surface to that of the exemplar, while keeping the reflectance layer from changing. 

\begin{figure*}[t]
	\begin{center}
		\begin{minipage}{\textwidth}
			\centering
			\subfloat[Input]
			{\includegraphics[width=0.245\textwidth]{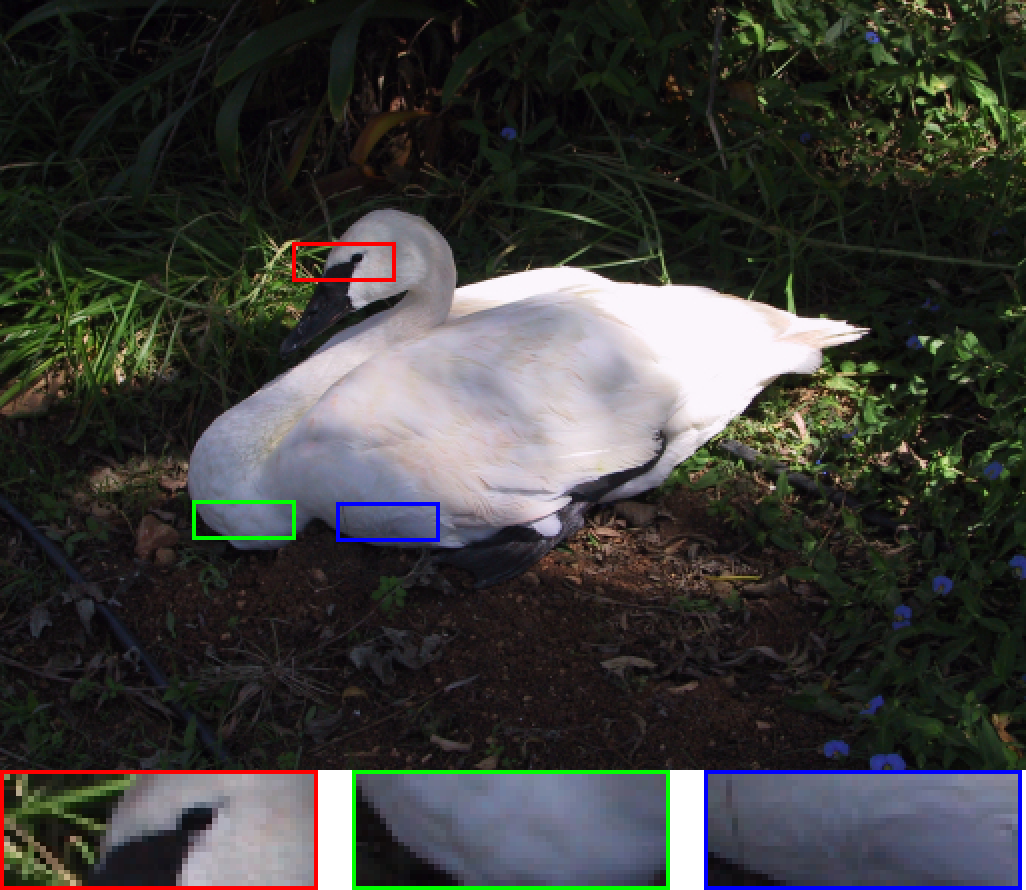}\label{fig:fig3a}}\hfil
			\subfloat[Exemplar (CLAHE)]
			{\includegraphics[width=0.245\textwidth]{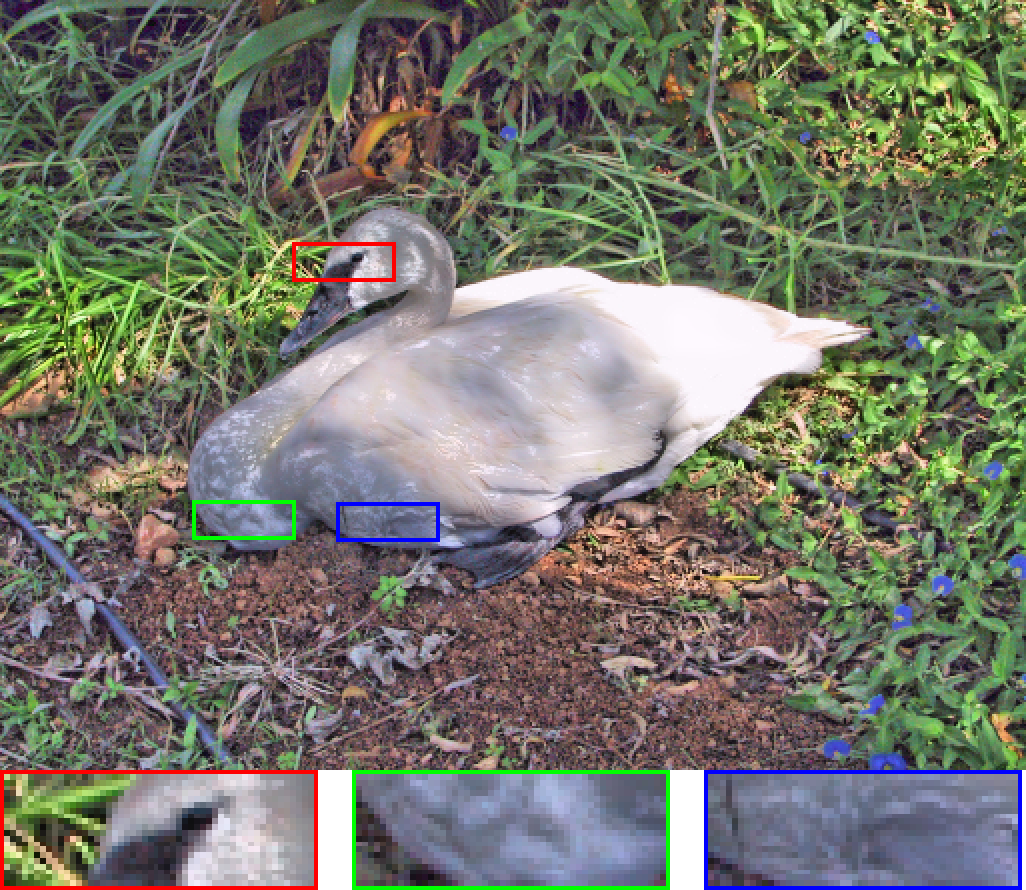}\label{fig:fig3b}}\hfil
			\subfloat[IIT+ GF (Ours)]
			{\includegraphics[width=0.245\textwidth]{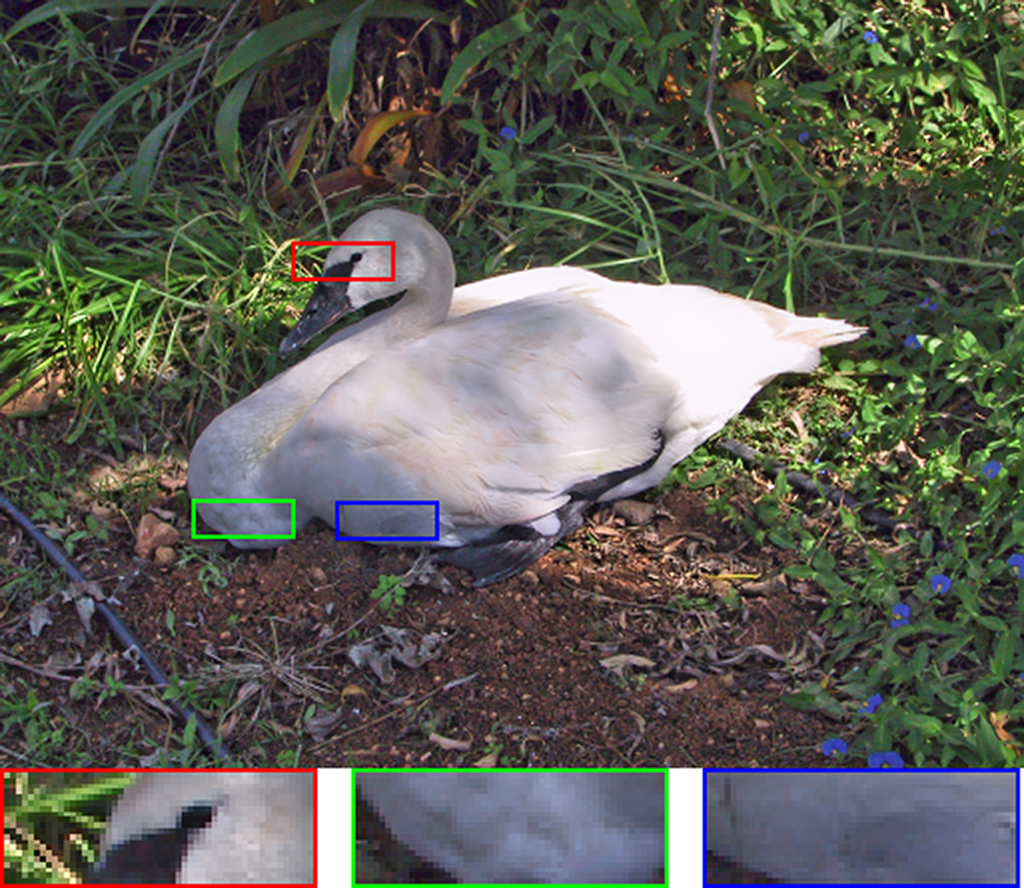}\label{fig:fig3c}}\hfil
			\subfloat[IIT+ BF (Ours) ]
			{\includegraphics[width=0.245\textwidth]{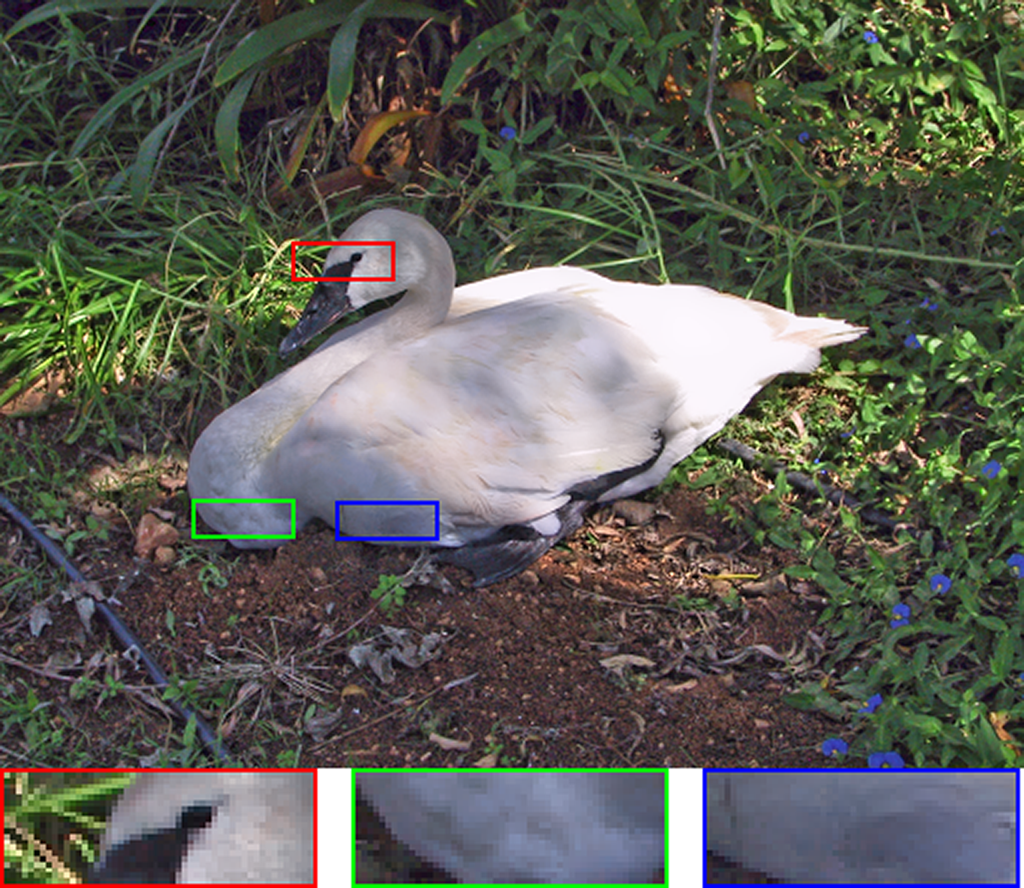}\label{fig:fig3d}}\hfil
		\end{minipage}
	\end{center}
	\caption{Visual results of our IIT algorithm with the Gaussian and bilateral filters, respectively. (a) Source, (b) CLAHE exemplar\cite{Zuiderveld1994Contrast}, (c) and (d) Our results. The noise and distortions are suppressed significantly in comparison to the CLAHE exemplar. TMQI (b)$\scriptsize{\sim}$(d): 0.845, 0.891, 0.898.} \label{fig:fig3}
\end{figure*}

\textbf{Content loss:} \quad We additionally introduce a so-called content loss to avoid the illumination over-fitting problem, which may occur when the pre-computed examplar has a strong over-stretched illumination. The content loss, in such cases, helps to give a remedy for the output global illumination. Specifically, we define the content loss ${E^c(\boldsymbol{o})}$ as,
\begin{equation}
	\label{eq:eq11}
	\begin{aligned}
		E^c(\boldsymbol{o})&={\sum_{i} {({o}_{i}-{s}_{i})}^2}.
	\end{aligned}
\end{equation}

We can verify that Eq. (11) is essentially defined on the illumination layers\footnote{{\small We also have ${E^c(\boldsymbol{o})\!=\!{\sum_{i} {({o}_{i}^{l}\!-\!{o}_{i}^{r}+{s}_{i}^{l}\!-\!{s}_{i}^{r})}^2}\!=\!{\sum_{i} {({o}_{i}^{l}-{s}_{i}^{l})}^2}}$} with ${o}_{i}^{r} \!=\! {s}_{i}^{r}$ under the illumination-invariant property of the reflectance layer.}. It is valuable to note that ${E^c(\boldsymbol{o})}$ is not always necessary but plays an auxiliary role to prevent output illumination from being over-dependent on exemplars. 

Until now, we have defined three photo-realistic losses to regularize the intrinsic images under the interpretation Eq. \ref{eq:eq1}. All of them are firstly derived from the intrinsic layers and then reduced on images under the well-known spatial-varying illumination and illumination-invariant reflectance assumption. This simplification helps to avoid the notorious intrinsic image decomposition and can be beneficial for many image illumination-related tasks, as they can be viewed as a local image translation between two illumination surfaces. Under this mild assumption, we declare that our IIT model can be directly used for these problems and significantly simplifies the scheme due to the unnecessary of intrinsic image decomposition.

However, the illumination loss in Eq. \ref{eq:eq5} requires an extra exemplar $\boldsymbol{c}$ for illumination guidance. In many illumination-related tasks, the ease of producing a suitable ``exemplar'' mainly comes from two aspects: (1) the exemplar $\boldsymbol{c}$ is not necessary to have high-quality details, because the majority of high-frequency details in $\boldsymbol{c}$ would be filtered out by the smoothing kernel ${\mathcal{K}}$;  and (2) a further correction in the reflectance loss will be carried out to remedy the remaining non-consistent structures, even a slight illumination dependency still exist after the smoothing process. It is worth noting that either illumination loss or reflectance loss can not independently provide a high-quality result and they play a complementary role together in regularizing the corresponding intrinsic layers. As a result, it is possible to produce an image with fine-balanced illumination and use it as an exemplar without paying too much attention to the accuracy of local details. We show that it is easy for many existing methods to generate a suitable examplar, which will be verified by a series of results in experiments.

\subsection{Optimization}
\label{optimization}

Now, we combine three loss functions and rewrite Eq. (2) in a matrix form:
\begin{equation}
	\label{eq:eq12}
	\begin{aligned}
		E(\boldsymbol{o}) =&\alpha {{\left\Vert \boldsymbol{K}^o\boldsymbol{o}-\boldsymbol{K}^c\boldsymbol{c}\right\Vert }_2^2}+\beta{{\left\Vert \boldsymbol{M}\boldsymbol{o} \right\Vert }_2^2}+\gamma{{\left\Vert \boldsymbol{o-s}\right\Vert }_2^2}, \\
		&\text{s.t.} \quad {\omega_{i,j}^{o}}= {\omega_{i,j}^{s}}, \quad {s_i}= {\sum_{j \in \Omega_{i}}{\omega_{i,j}^{s}{s_j}}},
	\end{aligned}
\end{equation}
where ${\boldsymbol{K}^{o}(\boldsymbol{K}^{c})}$ is a kernel affinity matrix, whose ${(i,j)}$th entry is ${\mathcal{K}_{i,j}^{o(c)}}$; and ${\boldsymbol{M}=[\boldsymbol{I}-\boldsymbol{W}]}$ is a sparse coefficient matrix with identity ${\boldsymbol{I}}$ and weight ${\boldsymbol{W}}$ containing entries ${\omega_{i,j}^{o}}$. Once ${\boldsymbol{K}^o}$, ${\boldsymbol{K}^c}$ and ${\boldsymbol{W}}$ are pre-computed, the output image can be reconstructed by solving the Eq. \ref{eq:eq12} directly.

We first compute the filter kernel ${\mathcal{K}}$. The simplest case can be the ${2}$-D Gaussian filter (GF), where ${\boldsymbol{f}}$ relies on pixel's position: ${\boldsymbol{p}=[p_x, p_y]}$ with the ${x}$ and ${y}$ directional coordinates ${p_{x}}$ and ${p_{y}}$ respectively. One can use bilateral filter (BF)\cite{tomasi1998bilateral} for more robust results, which considers both the pixel's position ${\boldsymbol{p}=[p_x, p_y]}$ and color intensities ${(R_i,G_i,B_i)}$.   
We set ${\delta_{\boldsymbol{f}_i}=\delta_s}$ for the Gaussian filter and  ${\delta_{\boldsymbol{f}_i}=(\delta_s,\delta_r)}$ for the bilateral filter. For simplicity, we set ${\boldsymbol{K} =\boldsymbol{K}^o=\boldsymbol{K}^c}$ in our experiments due to the unavailable of the image $\boldsymbol{o}$ in advance.

For the LLE weights, it may be unstable to take a direct solver\cite{roweis2000nonlinear}, when the number of neighbors of each point is larger than the space dimension. As described in\cite{saul2003think}, a remedy ${\boldsymbol{W}}$ can be computed by solving the point-wise regularized problem,
\begin{equation}
	\label{eq:eq13}
	\begin{aligned}
		\min{{{\sum_{i}({s_i}-{\sum_{j \in \Omega_{i}}{\omega_{i,j}^{s}{s_j}}}})^2}+\epsilon {\Vert \boldsymbol{\omega}^s\Vert}_2^2 }, \text{s.t.} \sum_{j \in \Omega_{i}} {\omega_{i,j}^{s}}=1.
	\end{aligned}
\end{equation}
 
We refer the reader to some more complex regularizers such as the modified LLE algorithm\cite{zhang2007mlle}, in which more robust solutions are given to distribute the contribution of neighbours to each point more uniformly.

\begin{algorithm}
	\caption{intrinsic image transfer (IIT)}
	\begin{algorithmic}[1]
		\renewcommand{\algorithmicrequire}{\textbf{Input:}}
		\renewcommand{\algorithmicensure}{\textbf{Output:}}
		\REQUIRE Images ${\{{s}_{i}\}_{i=1, \cdots, N}}$, ${\{{c}_{i}\}_{i=1, \cdots, N}}$ and ${\alpha, \beta, \gamma}$;
		\ENSURE Image ${\{{o}_{i}\}_{i=1, \cdots, N}}$;
		\STATE \textbf{Identifying filters:}
			\begin{enumerate}\setlength{\itemsep}{-\itemsep}
				\item Set parameters: ${\mathcal{N}_{i}, \delta_s, \delta_r}$ or ${(\mathcal{N}_{i}, \delta_s)};$
				\item Compute ${\mathcal{K}_{i,j}}$ in Eq. (4) and ${\boldsymbol{K}^o}$, ${\boldsymbol{K}^c}$ in Eq. \ref{eq:eq12};
			\end{enumerate}
		\STATE  \textbf{Computing} LLE \textbf{weights:} ${ \boldsymbol{W}}$
		\begin{enumerate}\setlength{\itemsep}{-\itemsep}
			\item Set parameters: ${\Omega_i, \epsilon}$;
			\item Find neighbors ${\Omega_i}$ for each pixel ${i}$;
			\item Compute ${\omega_{i,j}^s}$ in Eq. \ref{eq:eq13};
			\item Set ${\omega_{i,j}=\omega_{i,j}^s}$, and ${\boldsymbol{M=I-W}}$;
		\end{enumerate}
		\STATE  \textbf{Reconstruction}
		\begin{enumerate}\setlength{\itemsep}{-\itemsep}
			\item Compute the \emph{Laplacian matrix} ${\boldsymbol{L}}$ in Eq. \ref{eq:eq14};
			\item Solve Eq. \ref{eq:eq14} with the PCG algorithm;
		\end{enumerate}
	\end{algorithmic}
	\label{alg:algorithm1} 
\end{algorithm}

After obtaining the ${\boldsymbol{K}}$ and ${\boldsymbol{W}}$, Eq. \ref{eq:eq12} can be solved by setting ${dE/d\boldsymbol{o}=0}$, giving the following linear system:
\begin{equation}
	\label{eq:eq14}
	\begin{aligned}
		{(\alpha\boldsymbol{K}^{T} \boldsymbol{K} + \beta\boldsymbol{M}^{T} \boldsymbol{M} +\gamma \boldsymbol{I}) \boldsymbol{o}}={\alpha \boldsymbol{K}^T \boldsymbol{K} \boldsymbol{c} + \gamma \boldsymbol{s}},
	\end{aligned}
\end{equation}
where ${\boldsymbol{L} \!= \!\alpha\boldsymbol{K}^T\boldsymbol{K} \!+\! \beta\boldsymbol{M}^T\boldsymbol{M}\! +\!\gamma\boldsymbol{I}}$ is a large-scale and sparse \emph{Laplacian matrix}. Since ${\boldsymbol{L}}$ is symmetric and semi-positive, Eq. (14) can be solved with the solvers such as Gauss-Seidel method and preconditioned conjugate gradients (PCG) method\cite{saad2003iterative}. 

In summary, our IIT algorithm includes: identifying the filter kernel for illumination fitting, computing LLE weights to encode image reflectance, and embedding the reflectance layer for image reconstruction. The whole minimization scheme is presented in \textbf{Algorithm} \ref{alg:algorithm1}.

\section{Experimental Results}\label{experimental_results}  

In this section, we extensively illustrate the proposed IIT algorithm and show its many benefits to several illumination-related tasks, for example, illumination compensation, image enhancement, tone mapping, HDR image compression, and so on. For this purpose, we first explain how to generate a suitable ``exemplar'' and show its robustness under the proposed IIT framework. The performance is then presented under a series of numerical experiments and quantitative evaluations on natural image datasets. Our Matlab implementation runs on a desktop PC with an Intel i7 3.40 GHz, 32GB RAM, Win64, and it takes roughly 3 seconds to process a color image with $600 \times 400$ resolution. The code is also available: \url{https://github.com/QingXin96/Intrinsic_image_transfer}.

\subsection{Verification and Robustness}

We now verify the proposed IIT model with the aid of an extra ``exemplar'' image. As explained in Section 3.2,  our model may require a latent illumination layer in the sense that the underlying target illumination in many illumination-related tasks could deviate greatly from that of the original image. Nevertheless, it is not easy to provide such an illumination layer because of the inherent intractability of intrinsic image decomposition. Alternatively, we introduce a pre-computed ``exemplar'' and use it to provide the underlying illumination approximately. We have also interpreted that it is always tractable in practice to find a suitable exemplar under the well-known spatial-smoothing illumination and illumination-invariant reflectance prior knowledge. 

We take two representative exemplars into account to verify the easy-fulfilled requirements of the exemplar and its benefits to image illumination manipulation.  The first class of exemplars is generated by the simple TMO methods --- for example, contrast limited adaptive histogram equalization (CLAHE) method\cite{Zuiderveld1994Contrast} which provides an exemplar with fine-balanced illumination distribution but distorted local details. The second class of exemplars is given by the existing cutting-edge methods, which usually provide an exemplar exhibiting better quality with not so strong local artifacts as the CLAHE ones. They are selected in consideration of the different properties in color, saturation, textures, noises, and illumination conditions. Notice that, we specify the fine-balanced illumination distribution of the exemplar, but the proposed IIT framework does not rely on any specified exemplars.

Firstly, we show the results with a CLAHE exemplar.  We set ${\{ \mathcal{N}_{i}, \delta_s \} = \{5\times5, 2.0}\}$  and ${\{ \mathcal{N}_{i}, \delta_s,\delta_r \} = \{5\times5,}$ ${2.0, 0.2\}}$ for the Gaussian and  bilateral filters, respectively. In regard to the LLE weights, we choose ${\Omega_{i}=\{5\times5\}}$ and ${\epsilon=1e-5}$. The global parameters ${\alpha}$, ${\beta}$, ${\gamma}$ are set to 0.8, 100 and 0.2 respectively. We impose ${\alpha + \gamma =1}$ to limit the output result between the source and exemplar; and a large ${\beta}$ is utilized to force the output local textures to be more consistent with the source image. A larger $\gamma$ gives the result more identical to the source image. We adopt the default parameter-settings for the CLAHE algorithm\cite{Zuiderveld1994Contrast} without the specifications.  As shown in Fig.  \ref{fig:fig3},  the exemplar in Fig.  \ref{fig:fig3b} is generated by the CLAHE algorithm with the fine-balanced illumination but distorted local details, while our IIT algorithm exhibits significant improvements in suppressing the local noises and distortions, especially around the swan's ``neck'' and ``wing'' regions in Fig.  \ref{fig:fig3c} and Fig.  \ref{fig:fig3d}.  Moreover, tiny visual differences occur between using the Gaussian and bilateral filters.

\begin{figure*}[!t]
		\begin{center}	
		\begin{tabular}{c}
			\raggedright
			\begin{minipage}[t]{0.182\textwidth}
				{\includegraphics[width =\textwidth]{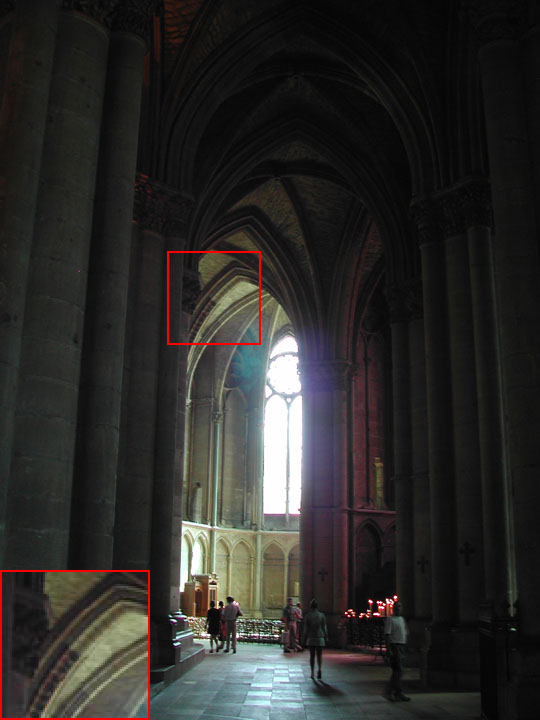}}\hfil
				\subfloat[Source]
				{\includegraphics[width =\textwidth]{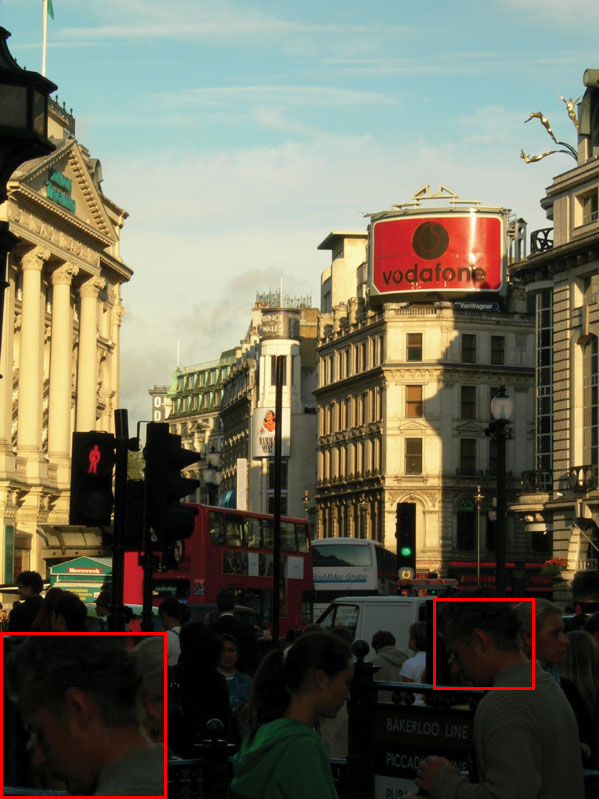}\label{fig:fig4a}}
			\end{minipage}
		\end{tabular}
		\begin{tabular}{|c}
			\centering
			\begin{minipage}[t]{0.76\textwidth}
				{\includegraphics[width = 0.24\textwidth]{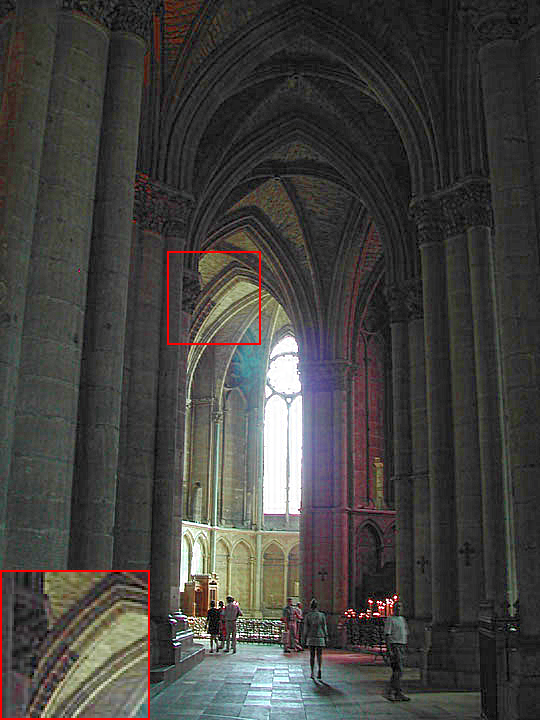}}\hfil
				{\includegraphics[width = 0.24\textwidth]{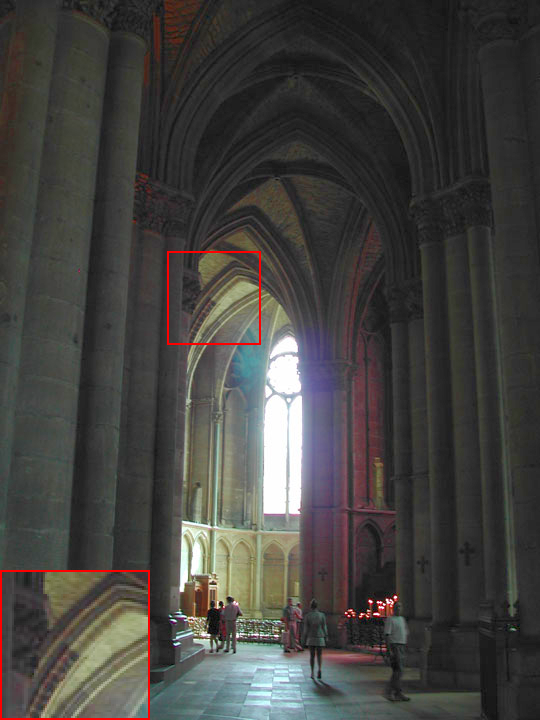}}\hfil
				{\includegraphics[width = 0.24\textwidth]{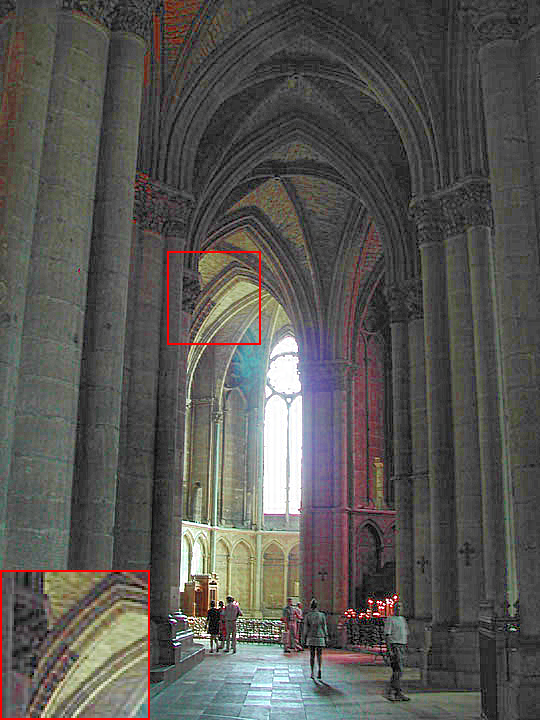}}\hfil
				{\includegraphics[width = 0.24\textwidth]{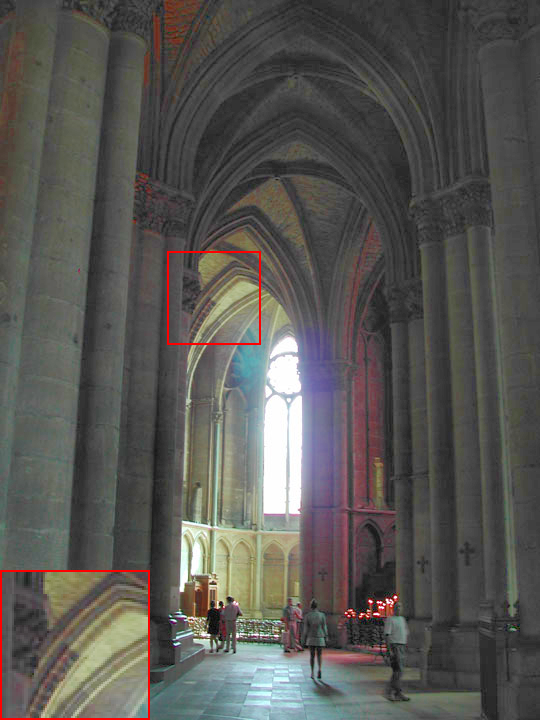}}\hfil
				\subfloat[Exemplar 1]
				{\includegraphics[width = 0.24\textwidth]{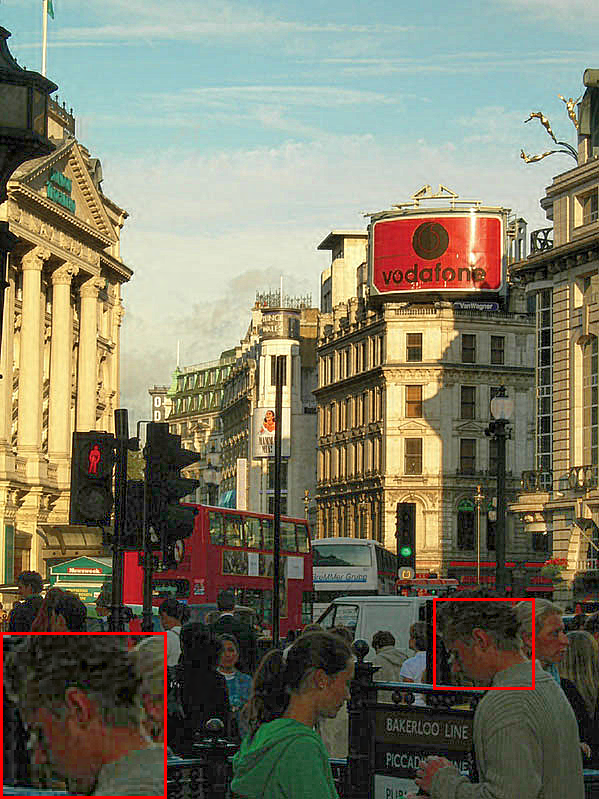}\label{fig:fig4b}}\hfil
				\subfloat[IIT + GF (Ours)]
				{\includegraphics[width = 0.24\textwidth]{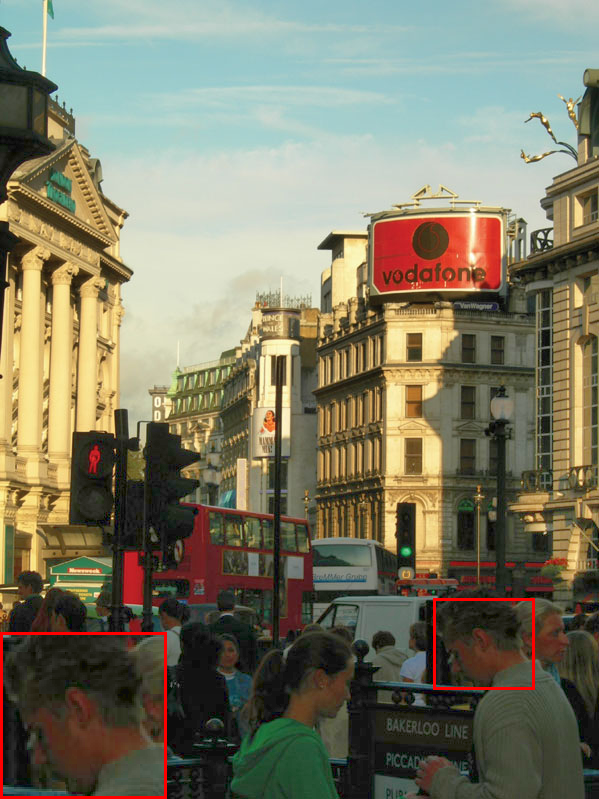}\label{fig:fig4c}}\hfil
				\subfloat[Exemplar 2]
				{\includegraphics[width = 0.24\textwidth]{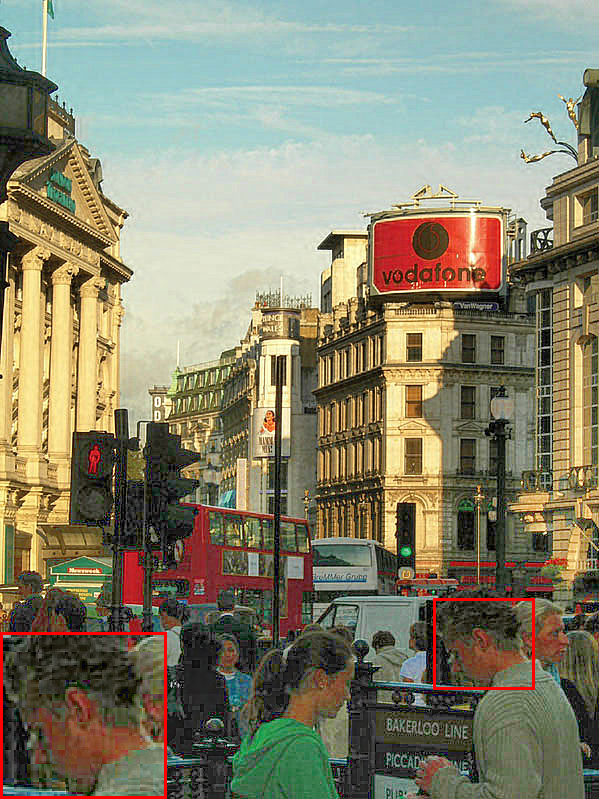}\label{fig:fig4d}}\hfil
				\subfloat[IIT + GF (Ours)]
				{\includegraphics[width = 0.24\textwidth]{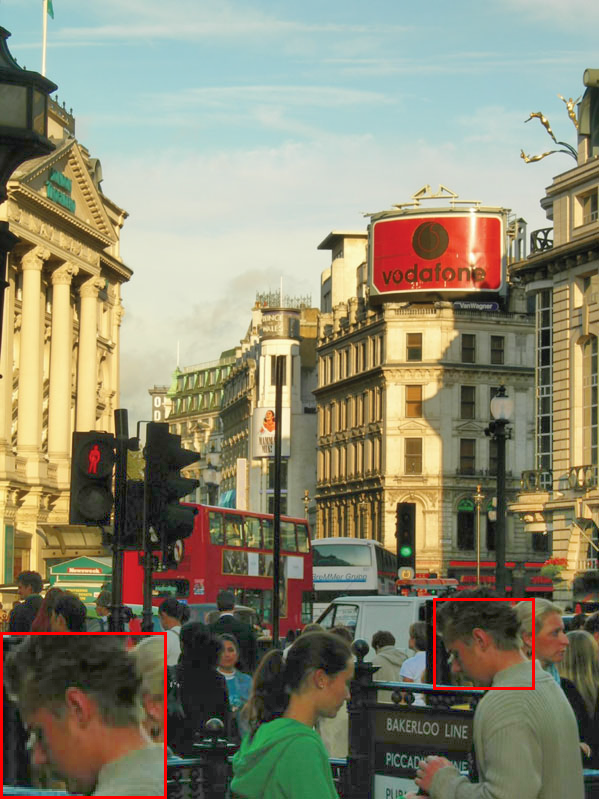}\label{fig:fig4e}}\hfil
			\end{minipage}
		\end{tabular}
		\end{center}
	\caption{Visual comparison of our IIT algorithm by using exemplars with different levels of brightness. (a) Input image; (b) and (d) CLAHE exemplars\cite{Zuiderveld1994Contrast}; (c) and (e) our IIT results.} 
	\label{fig:fig4}
\end{figure*}

\begin{figure}[!b]
	\begin{center}	
		\begin{minipage}{0.49\textwidth}
			\centering
			{\includegraphics[width=0.19\textwidth]{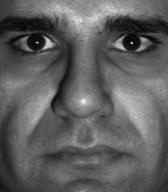}}\hfil
			{\includegraphics[width=0.19\textwidth]{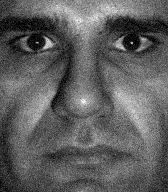}}\hfil
			{\includegraphics[width=0.19\textwidth]{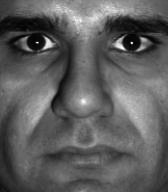}}\hfil
			{\includegraphics[width=0.19\textwidth]{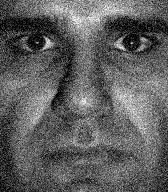}}\hfil
			{\includegraphics[width=0.19\textwidth]{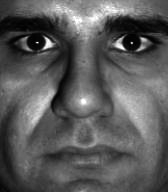}}\hfil
			\subfloat[]
			{\includegraphics[width=0.19\textwidth]{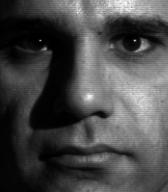}\label{fig:fig5a}}\hfil
			\subfloat[]
			{\includegraphics[width=0.19\textwidth]{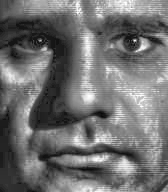}\label{fig:fig5b}}\hfil
			\subfloat[]
			{\includegraphics[width=0.19\textwidth]{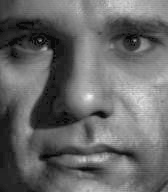}\label{fig:fig5c}}\hfil
			\subfloat[]
			{\includegraphics[width=0.19\textwidth]{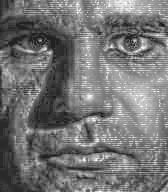}\label{fig:fig5d}}\hfil
			\subfloat[]
			{\includegraphics[width=0.19\textwidth]{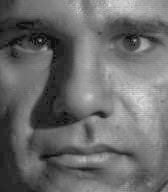}\label{fig:fig5e}}\hfil
		\end{minipage}
	\end{center}
	\caption{Illumination compensation on Yale Face dataset\cite{schroff2011pose}. (a) Source, (b) and (d) CLAHE exemplars\cite{Zuiderveld1994Contrast}, (c) and (e) results (IIT+ BF). TMQI, top: 0.935, 0.985, 0.833, 0.981; bottom: 0.922, 0.967, 0.897, 0.965.}
	\label{fig:fig5}
\end{figure} %

We further explain the ability of our IIT method in fitting the latent image illumination determined by the exemplars. As shown in Fig.  \ref{fig:fig4}, we present two typical exemplars given by the CLAHE algorithm\cite{Zuiderveld1994Contrast}. The exemplars in Fig.  \ref{fig:fig4b} and Fig.  \ref{fig:fig4d} should have different underlying illumination, as they exhibit different levels of brightness regardless of the slight texture distortions. In this case, we set $\alpha \!=\! 1.0$ and $\gamma \!= \!0$ to enforce the output be close to the exemplar.  It is noticeable that the outputs in Fig.  \ref{fig:fig4c} and Fig.  \ref{fig:fig4e} reveal identical illumination distribution as that of the exemplars, and the non-consistent details in the exemplars are significantly suppressed in both cases. The results prove that a balanced image illumination instead of high-quality local details is the key ingredient to our IIT method. This observation is consistent with the assumption that the exemplar plays a determining role in improving the balance of image illumination, while local textures have little contribution to the output illumination.

We additionally illustrate the robustness of our IIT method to the exemplar with strong local degraded details. As shown in Fig.  \ref{fig:fig5}, we interpret the illumination compensation results on Yale Face dataset\cite{schroff2011pose}. In this case, two typical exemplars (b) and (d) are introduced to show the robustness of our method to noise (Top) and textural distortions (Bottom). In the top row, the exemplars are cropped with different levels of Gaussian noise; while, in the bottom row, the exemplars suffer from serious textural distortions given by the CLAHE  algorithm\footnote{As interpreted in CLAHE\cite{Zuiderveld1994Contrast}, we set the ``NumTiles = ${16\times16}$'' and ``NumTiles = ${32\times32}$'' for the exemplars in Fig.  \ref{fig:fig5b} and Fig. \ref{fig:fig5d}, respectively. A larger ``NumTiles'' leads to stronger textural distortions.}.  Again, our IIT method produces the satisfying results Fig.  \ref{fig:fig5c} and Fig. \ref{fig:fig5e} that have identical illumination distribution as the exemplars, but the non-consistent textures are significantly suppressed, especially in ``face'' regions. Moreover, the quality of outputs only drops slightly if we only increase the level of textural distortions but keep the illumination distribution invariant.

The above experimental results imply that image illumination can be faithfully controlled with the aid of an ``exemplar'',  because there is no need to pay much attention to local distortions and artifacts. As a result, it is easy to obtain a suitable exemplar by using many existing methods.  This advantage is mainly beneficial from the use of the smoothing operator in illumination loss and the translation-invariant property of the LLE ``encoding'' model. On the one hand, the local details are mostly filtered out by the smoothing filter in the illumination loss, which remarkably weakens the impact of textural distortions; on the other hand, the inaccuracy of the smoothed illumination would be further corrected by the LLE ``encoding'' model. The two aspects help to penalize the non-consistent structures existing in the exemplar significantly, thereby giving a practical way to reconstruct natural-looking results with high-quality consistent local details.

\begin{figure*}[!hbt]
	\begin{center}
		\begin{minipage}{\textwidth}
			{\includegraphics[width=0.24\textwidth]{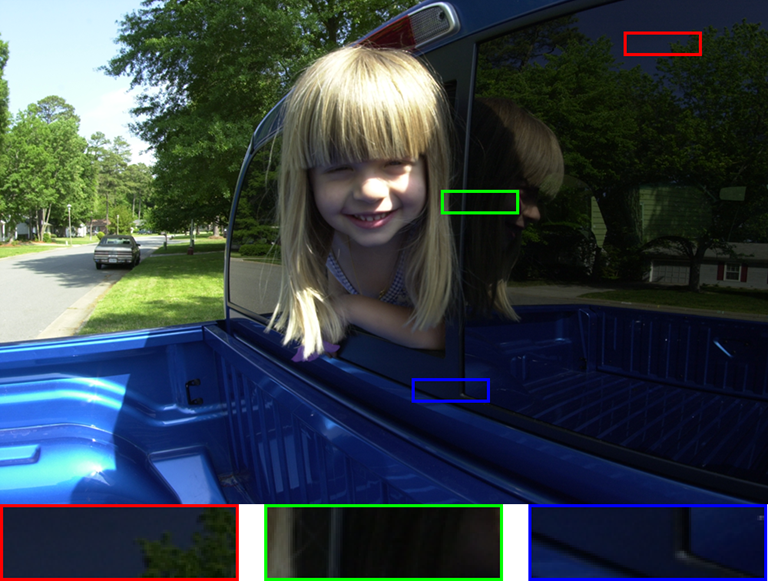}}\hfil
			{\includegraphics[width=0.24\textwidth]{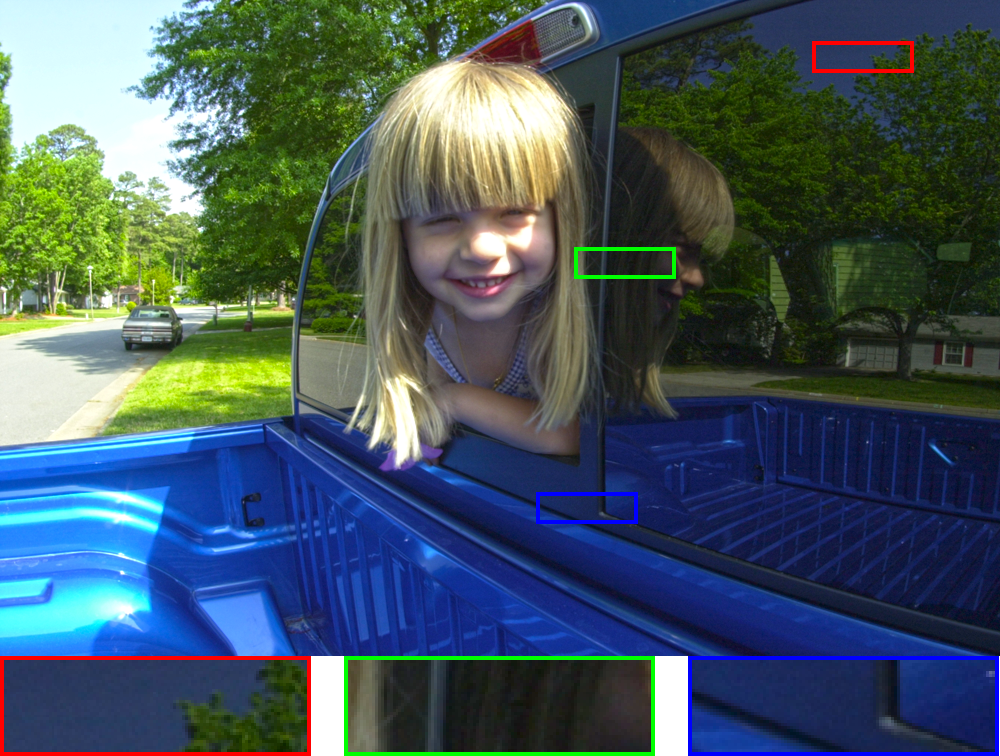}}\hfil
			{\includegraphics[width=0.24\textwidth]{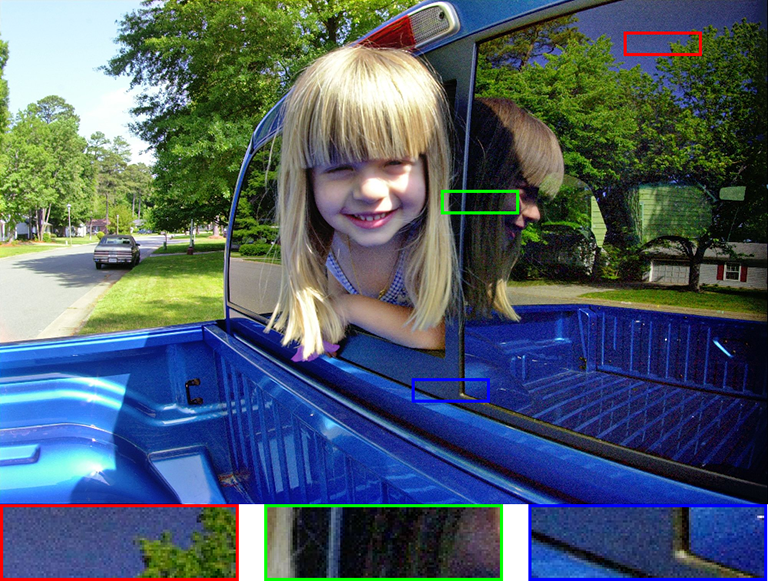}}\hfil
			{\includegraphics[width=0.24\textwidth]{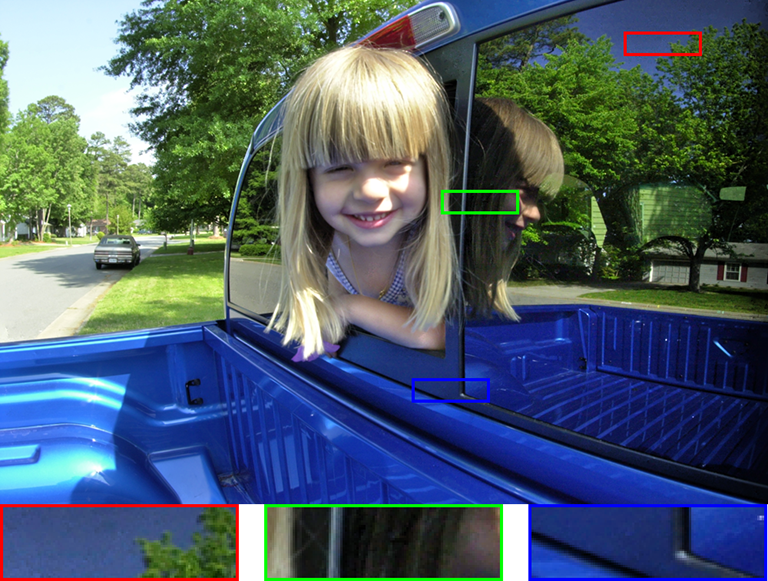}}\hfil
			\subfloat[Input]
			{\includegraphics[width=0.24\textwidth]{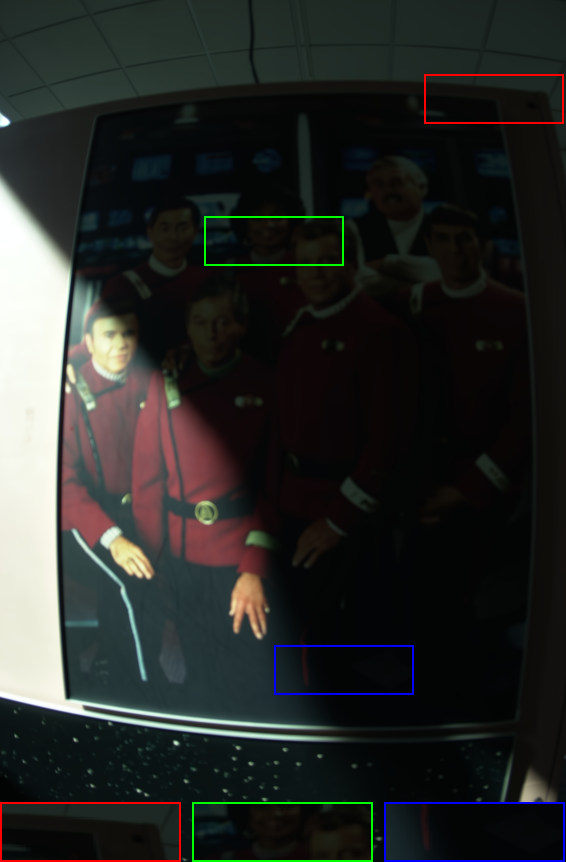}}\hfil
			\subfloat[Photoshop CC\cite{photoshop_cc} ]
			{\includegraphics[width=0.24\textwidth]{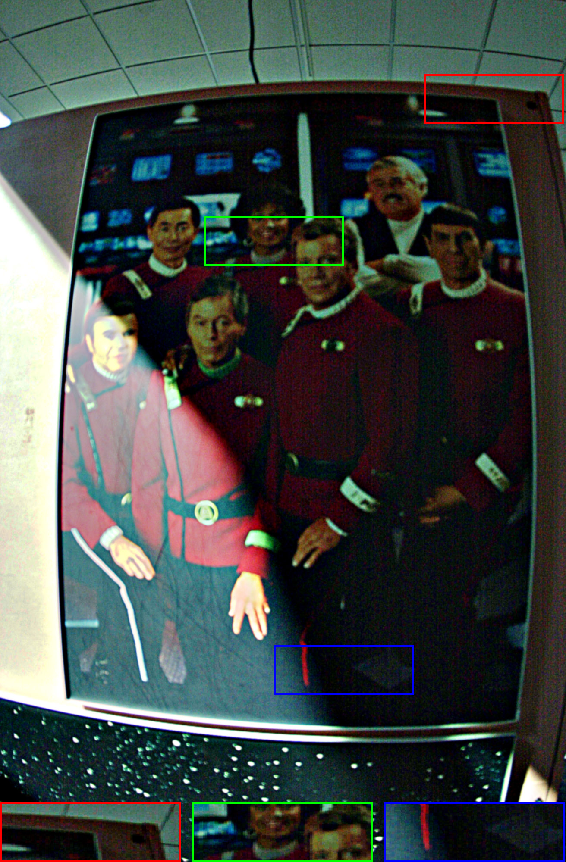}}\hfil
			\subfloat[NASA Retinex\cite{nasa_retinex}]
			{\includegraphics[width=0.24\textwidth]{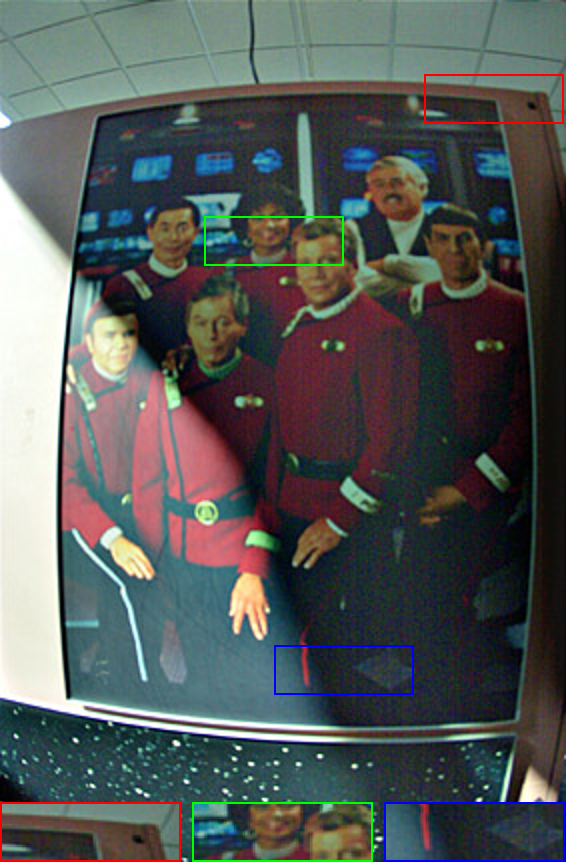}}\hfil
			\subfloat[IIT+ BF (Ours)]
			{\includegraphics[width=0.24\textwidth]{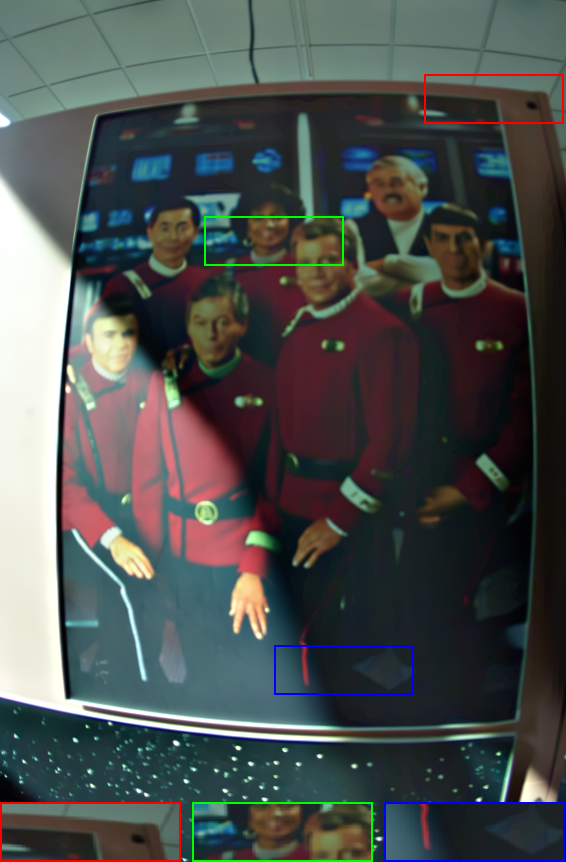}}\hfil
		\end{minipage}
	\end{center}
	\caption{Visual comparison of image enhancement. The exemplars in (c) are produced by the state-of-art NASA Retinex\cite{nasa_retinex}. TMQI,  top (b)$\scriptsize{\sim}$(d): 919, 0.943, 0.946; and bottom (b)$\scriptsize{\sim}$(d): 0.922, 0.916, 0.944.}
	\label{fig:fig6}
\end{figure*}

\begin{table*}[!ht]
	\scriptsize
	\caption{Quantitative evaluation on Cityscapes\cite{Cordts2016Cityscapes}, NASA Retinex\cite{nasa_retinex} and DPED\cite{ignatov2017dslr} datasets. The exemplars in the datasets are given by the WESPE, Retinex and CLAHE methods, respectively.}
	\begin{center}
		\setlength{\tabcolsep}{2.4mm}{
			\begin{tabular}{|c|c c c| c c c|c c c|c c c|}
				\hline
				&\multicolumn{3}{|c|}{Cityscapes (WESPE)}                     							 	 	& \multicolumn{3}{|c|}{NASA (Retinex)}					           						  & \multicolumn{3}{|c|}{DPED (CLAHE)} 							 	\\
				\hline
				Method		  &  (SF / SN) TMQI        							& IL-NIQE    	    & NIMA               	&  (SF / SN) TMQI           					& IL-NIQE    		& NAMA                & (SF / SN) TMQI                  		  & IL-NIQE 			& NIMA\\
				\hline
				NASA Retinex  & (- / -) -             						    & -                	& -         	        & (0.916 / \textbf{0.731}) \underline{0.937}     	& 20.71        		& \textbf{4.562}      & (- / -) -			    				  & - 					& -\\
				Photoshop CC  &(\textbf{0.988} / 0.323) 0.887       			    & 17.37          	& 3.859                 & (0.948 / 0.428) 0.892    		  				& 21.65       		& 4.003      		  & (\textbf{0.982} / 0.507) 0.916  		      & 22.38				& 4.479\\
				APE           &(0.946 / 0.272) 0.840       		  			        & 24.25       		& 4.002       		    &(0.981 / 0.618) 0.937    		      		    & \textbf{20.91}    &3.922                & (\underline{0.980} / 0.566) 0.927		      & 21.62				& 4.613\\
				Google Nik    & (0.927 / 0.527) 0.906       		  		        & 21.32       		& 4.131       		    &(0.968 / 0.812) 0.965			   				& 23.15        		&3.822                & (0.963 / 0.567) 0.925            	    	  &\underline{21.53}	& 4.523\\
				WESPE         &(0.915 / \textbf{0.839}) 0.956		 			    & 20.25        		& \textbf{4.338}        &(- / -) -               		 			    & -             	& -             	  & (0.931 / \textbf{0.626}) 0.928       		  & 22.25				& 4.534\\
				\hline
				IIT+GF (Ours) &(0.979 / \underline{0.835}) \textbf{0.971}     	    & 4.252          	& \underline{4.313}     &(\underline{0.957} / 0.650) 0.936 				& \underline{20.62} & \underline{4.475}   & (0.969 / 0.587) \underline{0.929}  		  & 21.95				& \textbf{4.555}\\
				IIT+BF (Ours) &(\underline{0.981} / 0.826) \underline{0.970}        & \textbf{16.48}    & 4.293                 & (\textbf{0.960} / \underline{0.678}) \textbf{0.942}  & \textbf{20.57}     & 4.470         & (0.973 / \underline{0.589}) \textbf{0.931}  &\textbf{21.31}		& \underline{4.540}\\
				\hline
		\end{tabular}}
	\end{center}
	\label{tab:tab1}
\end{table*}

Secondly, we validate our IIT algorithm with an exemplar given by the state-of-the-art. In general, they produce an exemplar with much better quality than that of the CLAHE method. We adopt the default configurations or preferable parameter-settings as suggested in these methods. We set ${\alpha, \beta}$ and ${\gamma}$ to 0.95, $[10, 100]$ and 0.05, respectively, and the parameter $\beta$ may vary according to the level of artifacts existing in exemplars. We let ${\alpha \gg \gamma}$ in view of the balanced illumination distribution of the exemplars. As shown in Fig.  \ref{fig:fig6}, we compare the results with the Photoshop CC\cite{photoshop_cc} and NASA Retinex\cite{nasa_retinex} methods, where the exemplars are provided by the cutting-edge NASA Retinex\cite{nasa_retinex}.  It is notable that Photoshop CC\cite{photoshop_cc} gives the results with limited improvement in dark regions based on the built-in ``HDR tone mapping'' tool, while the NASA Retinex method produces high-quality results with vivid color and contrast but also suffers from strong noise. In contrast, our IIT method attains similar enhanced results as the NASA Retinex with more consistent local textures compared with Photoshop CC\cite{photoshop_cc} and NASA Retinex\cite{nasa_retinex} results.

\begin{figure*}[!t]
	\begin{center}
		\begin{minipage}{\textwidth}
			\centering
			\subfloat[Input]
			{\includegraphics[width=0.245\textwidth]{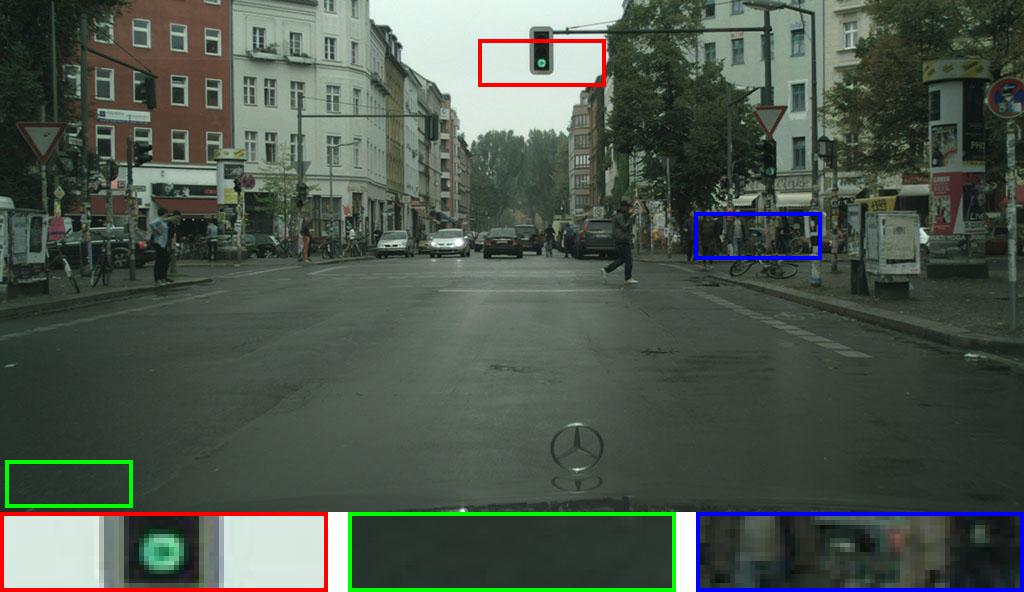}\label{fig:fig7a}}\hfil
			\subfloat[CLAHE\cite{Zuiderveld1994Contrast}]
			{\includegraphics[width=0.245\textwidth]{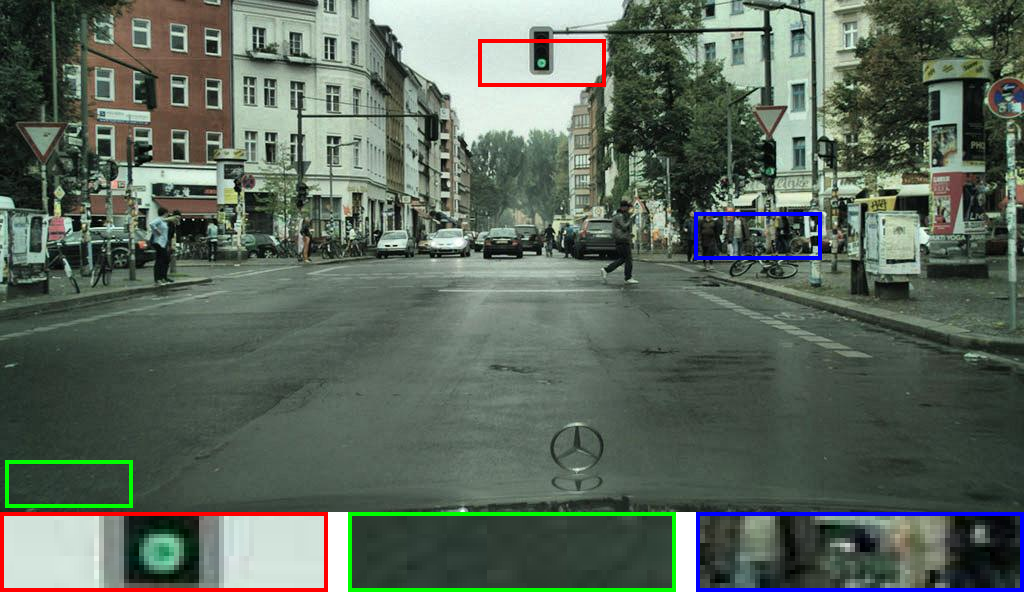}\label{fig:fig7b}}\hfil
			\subfloat[Photoshop CC \cite{photoshop_cc}]
			{\includegraphics[width=0.245\textwidth]{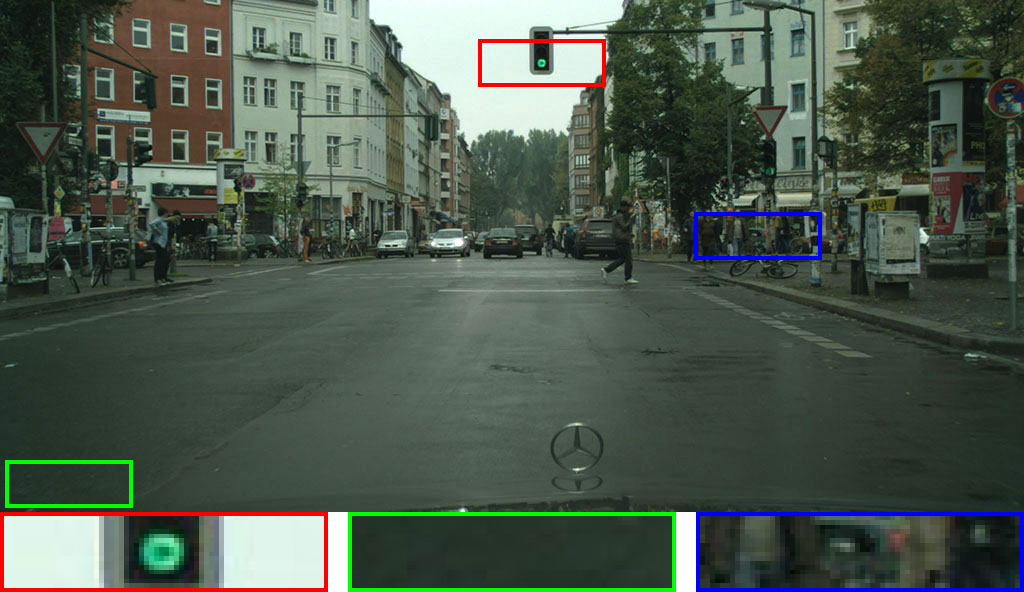}\label{fig:fig7c}}\hfil
			\subfloat[APE]
			{\includegraphics[width=0.245\textwidth]{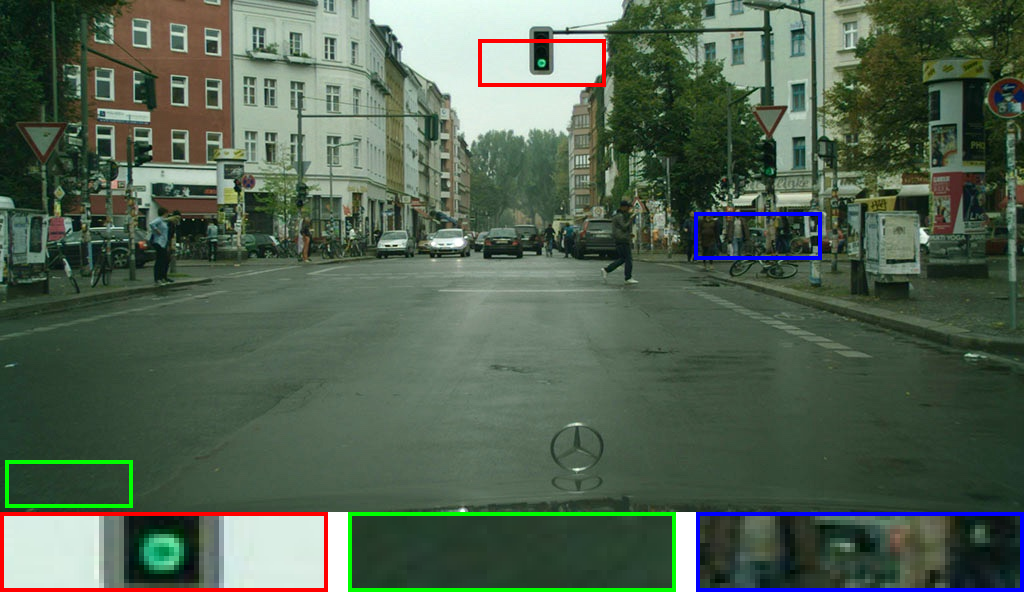}\label{fig:fig7d}}\hfil
			\subfloat[Google Nik\cite{google_nik}]
			{\includegraphics[width=0.245\textwidth]{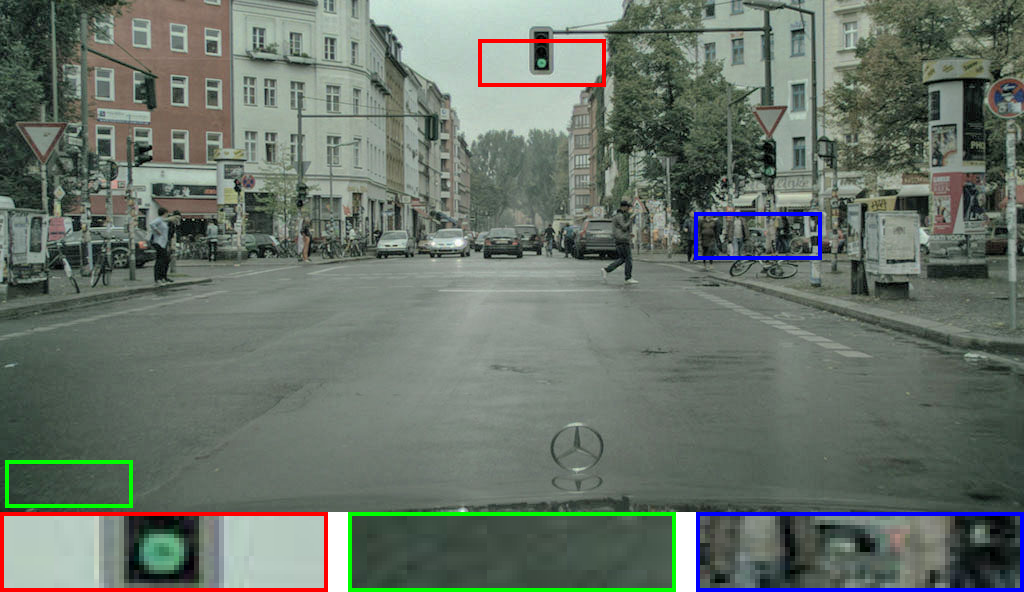}\label{fig:fig7e}}\hfil
			\subfloat[WESPE\cite{ignatov2017wespe}]
			{\includegraphics[width=0.245\textwidth]{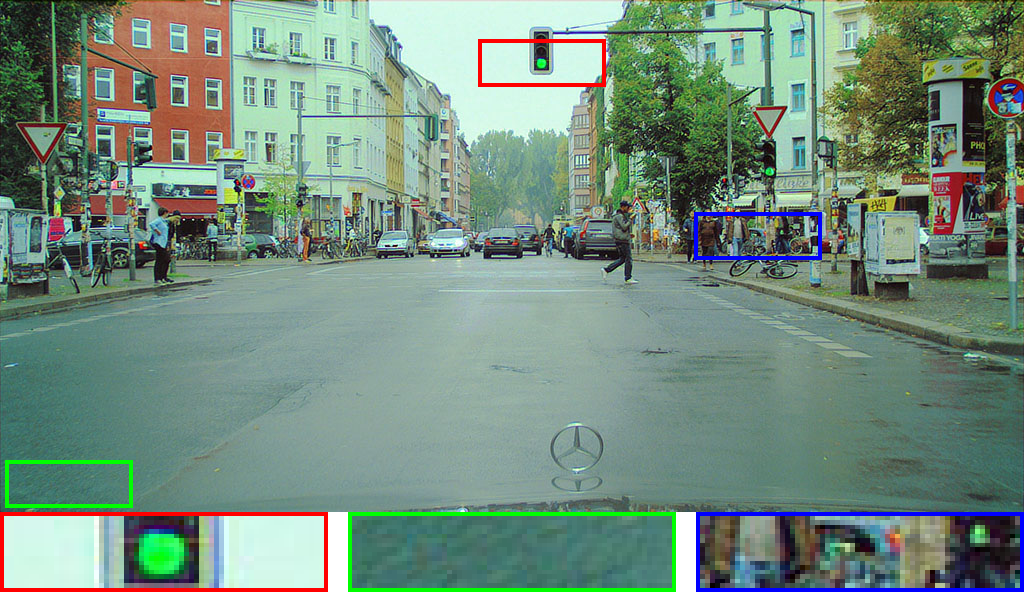}\label{fig:fig7f}}\hfil
			\subfloat[IIT+GF (Ours)]
			{\includegraphics[width=0.245\textwidth]{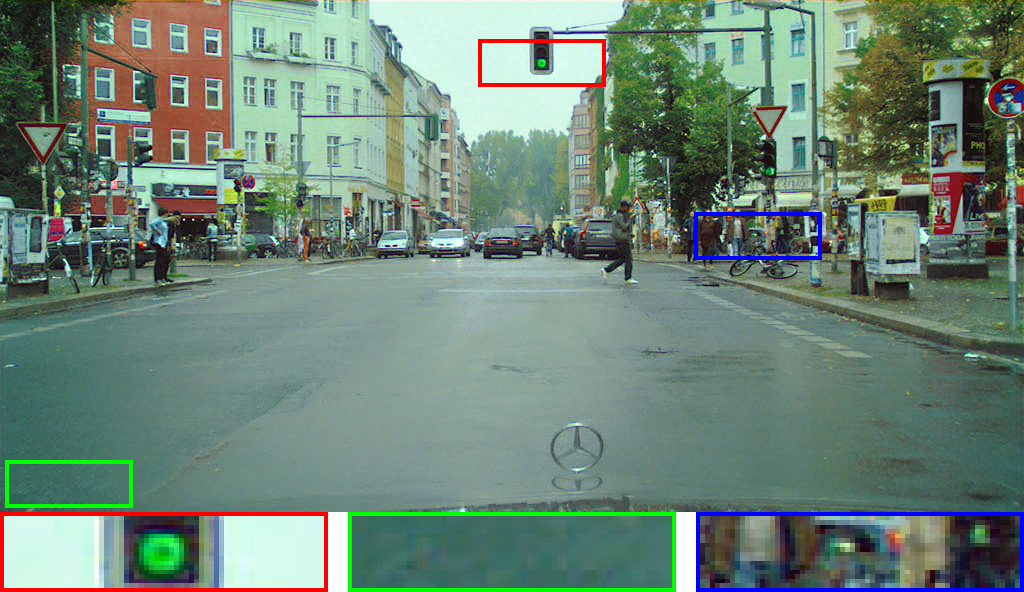}\label{fig:fig7g}}\hfil
			\subfloat[IIT+BF (Ours)]
			{\includegraphics[width=0.245\textwidth]{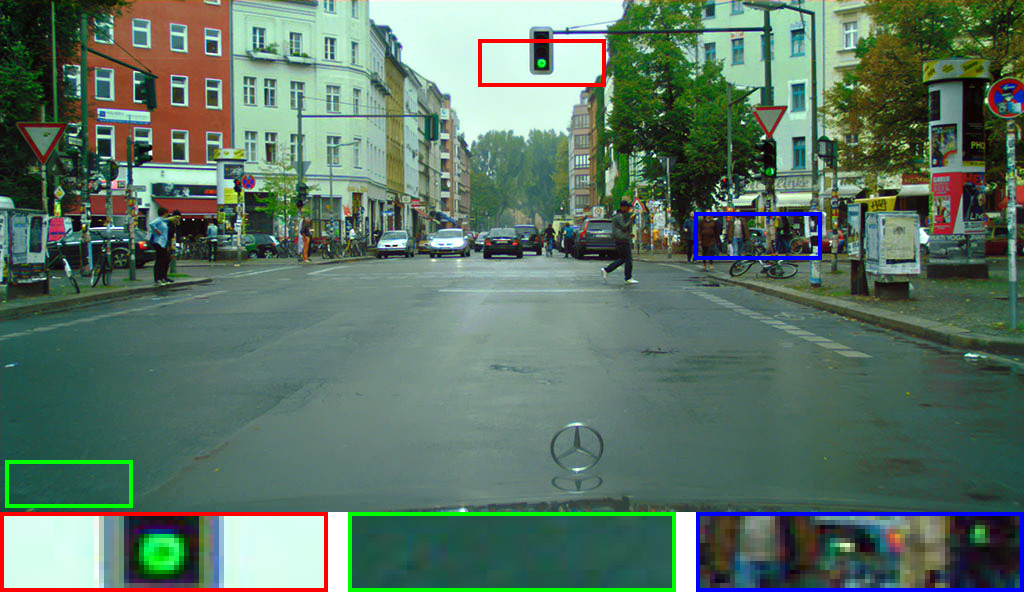}\label{fig:fig7h}}
		\end{minipage}
	\end{center}
	\caption{Visual comparison on Cityscapes dataset\cite{Cordts2016Cityscapes}.
		The exemplars used  in (g) and (h) are given by the state-of-art (f) WESPE\cite{ignatov2017wespe}. TMQI, (b)$\scriptsize{\sim}$(h): 0.891, 0.913, 0.973, 0.893, 0.801, 0.921, 0.942.}
	\label{fig:fig7}
\end{figure*}

The result is further demonstrated on the Cityscapes dataset\cite{Cordts2016Cityscapes}, in which the images suffer from visible degradation in color, saturation, and textures due to the low resolution and inappropriate illumination conditions. In this situation, it is often difficult for many existing methods to recover high-quality results. As shown in Fig.  \ref{fig:fig7}, the prevailing commercial software, including Google Nik (GN)\cite{google_nik}, Apple Photo Enhancer (APE) and Adobe Photoshop CC 2018\cite{photoshop_cc} also reveal drawbacks in producing natural-looking results, where the automatic parameter-settings are adopted for comparison. Recent deep-learning approaches such as WESPE\cite{ignatov2017wespe} provide a solution to visual-pleasant results, but there still exists some unpleasant artifacts such as over-exaggerated details around the salient edges. In contrast, our IIT method can be applied to correct these artifacts for high-quality consistent local details. As we can see in Fig.  \ref{fig:fig7g} and Fig.  \ref{fig:fig7h}, we take the WESPE result as the exemplar and the local structures of ``street lamp'' are precisely preserved and the ``road'' reveals little texture distortions compared with the other methods.

We now have investigated the role of exemplars in our IIT method and demonstrated the robustness to produce high-quality results under different types of exemplars. We also conclude that many existing methods can be used to provide such an exemplar. As interpreted before, the whole procedure, with the aid of an exemplar, is eventually formulated into a generalized optimization framework, giving a closed-form solution to a wide range of illumination-related tasks. The favorable results definitively benefit from both the smoothing operators in illumination loss --- which helps to reduce the impact of local details in the exemplar, and the LLE encode in reflectance loss --- which adaptively extracts local details from the original image and then embeds them onto a new smoothing surface given by the exemplar. The merit is the use of an exemplar, which significantly simplifies image illumination manipulation because of avoiding an explicit intrinsic image decomposition.

\begin{figure}[!b]
	\begin{center}
		\begin{minipage}{0.48\textwidth}
			\centering
			\subfloat[Ward's\cite{openexr_ward}]
			{\includegraphics[width=0.325\textwidth]{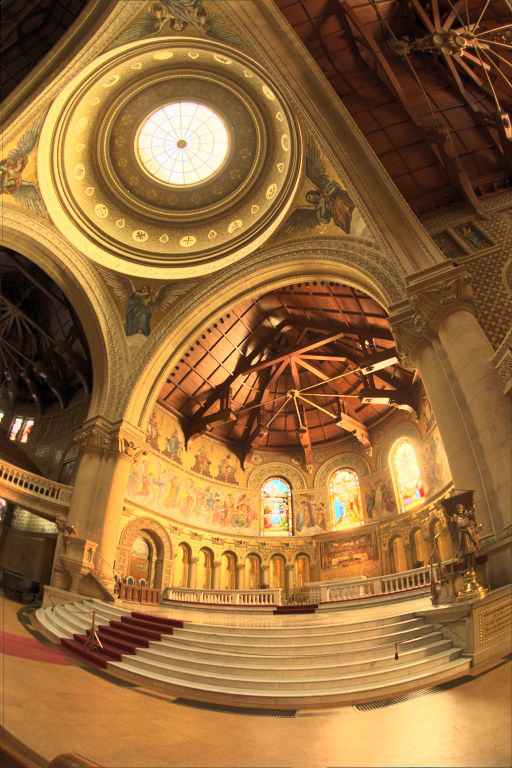}}\hfil
			\subfloat[GD\cite{fattal2002gradient}]
			{\includegraphics[width=0.325\textwidth]{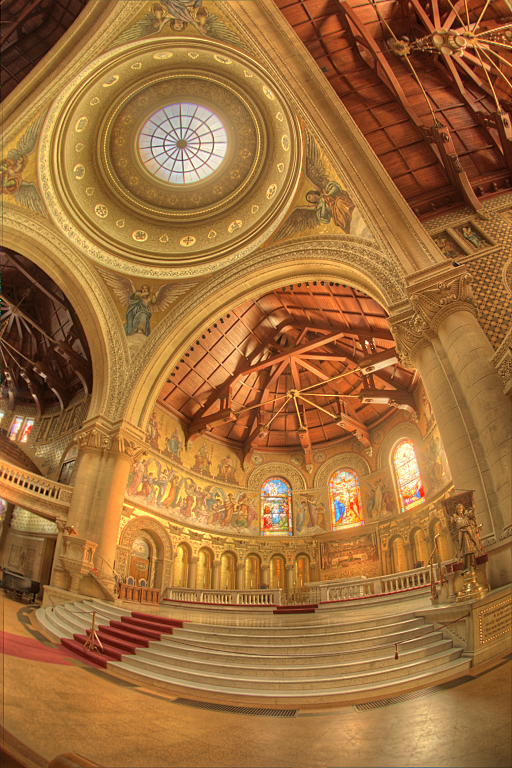}}\hfil
			\subfloat[IIT+BF (Ours)]
			{\includegraphics[width=0.325\textwidth]{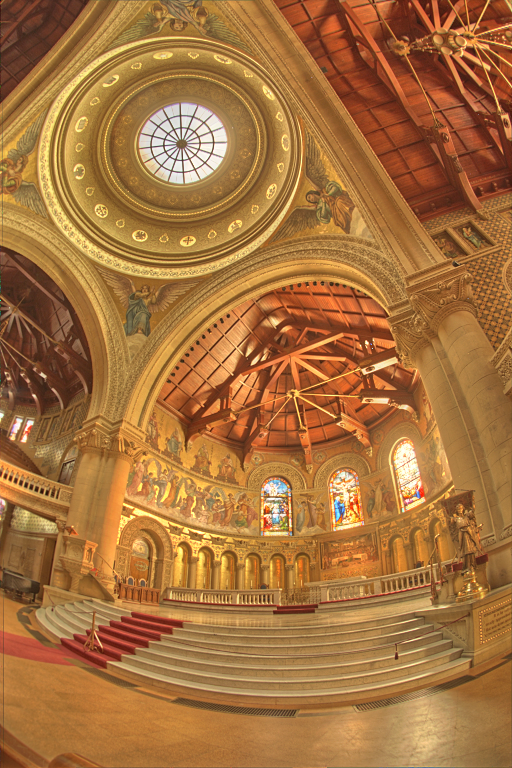}}\hfil
			\subfloat[Ward's\cite{openexr_ward}]
			{\includegraphics[width=0.325\textwidth]{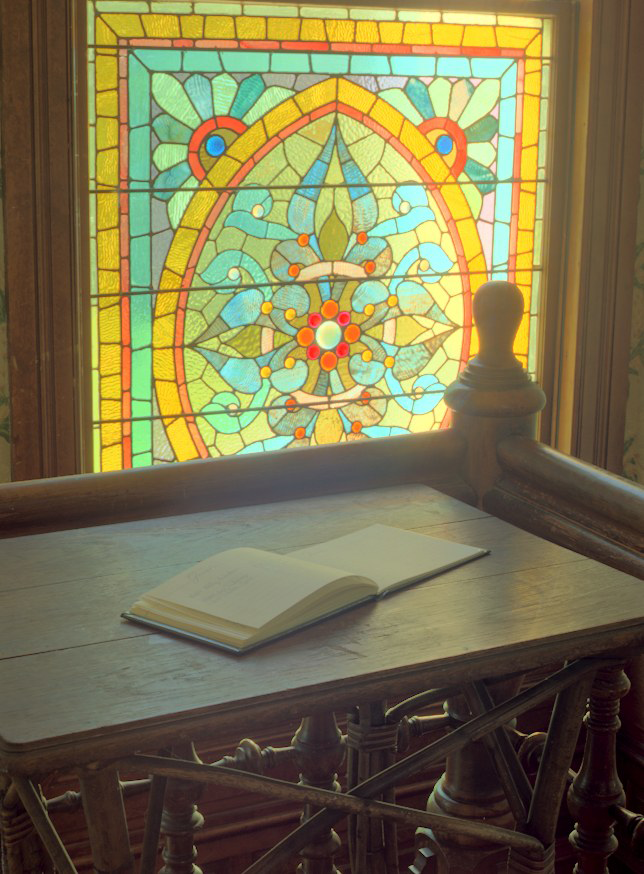}}\hfil
			\subfloat[WLS\cite{farbman2008edge}]
			{\includegraphics[width=0.325\textwidth]{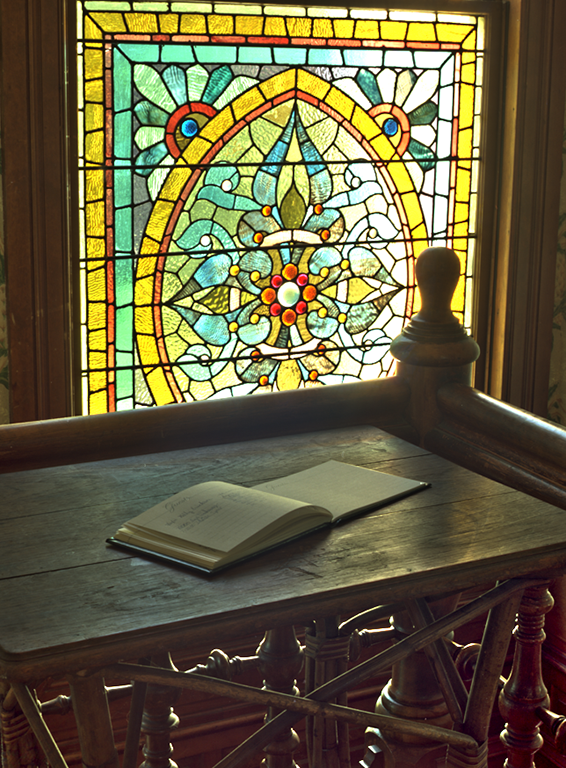}}\hfil
			\subfloat[IIT+BF (Ours)]
			{\includegraphics[width=0.325\textwidth]{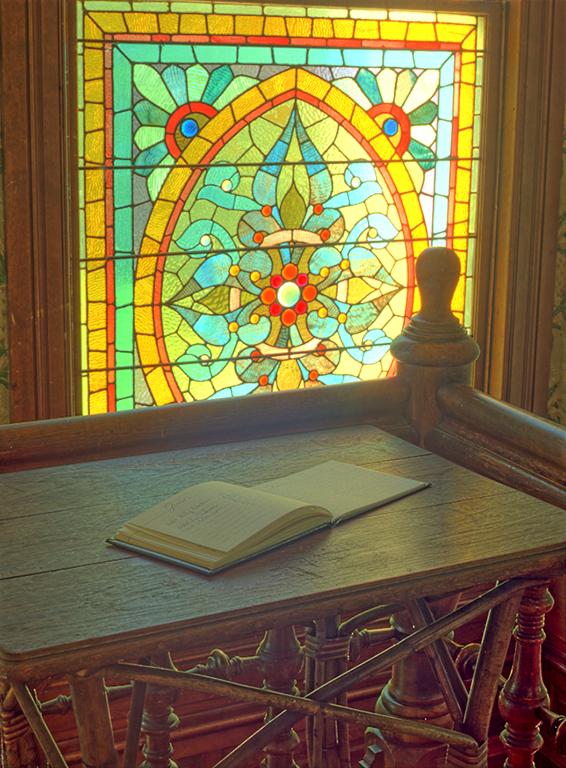}}\hfil
		\end{minipage}
	\end{center}
	\caption{Visual comparison of  the HDR image compression using the CLAHE exemplars\cite{Zuiderveld1994Contrast}. TMQI, top (a)$\scriptsize{\sim}$(c): 0.923, 0.948, 0.944; bottom (d)$\scriptsize{\sim}$(f): 0.912, 0.940, 0.953.}
	\label{fig:fig8}
\end{figure}

\subsection{Quantitative Evaluation}

In this section, we further take a quantitative evaluation based on the following datasets:  Cityscapes\cite{Cordts2016Cityscapes}, NASA Retinex\cite{nasa_retinex} and DPED\cite{ignatov2017dslr}. The Cityscapes\cite{Cordts2016Cityscapes} dataset has 20 low-resolution images with slightly unbalanced illumination distribution, which are challenging for many existing methods; while NASA Retinex\cite{nasa_retinex} and DPED\cite{ignatov2017dslr} datasets have 26 and 16 images respectively, which have higher resolution but also reveals strong imbalance illumination distribution. The proposed IIT method is compared with the cutting-edge methods, including NASA Retinex\cite{nasa_retinex}, Google Nik (GN)\cite{google_nik}, Apple Photo Enhancer (APE),  Adobe Photoshop CC 2018\cite{photoshop_cc}, WESPE method\cite{ignatov2017wespe}, and the exemplars in the datasets are provided by the Retinex\cite{nasa_retinex}, WESPE and CLAHE methods, respectively.
\begin{figure*}[!t]
	\begin{center}
		\begin{minipage}{\textwidth}
			\centering
			{\includegraphics[width=0.26\textwidth,height=0.14\textheight]{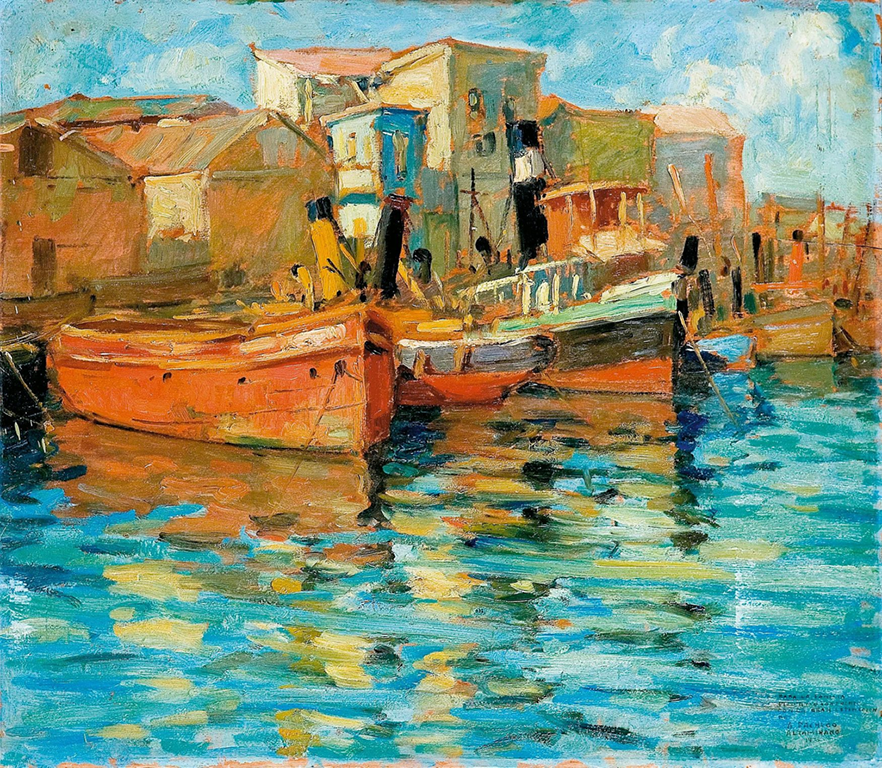}}\hfil
			{\includegraphics[width=0.26\textwidth,height=0.14\textheight]{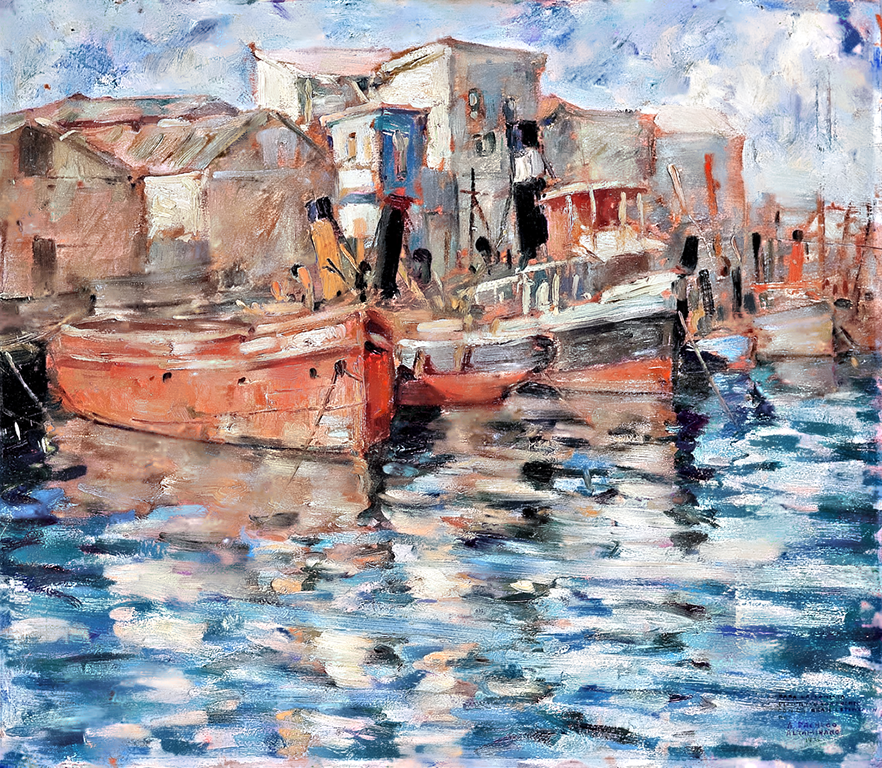}}\hfil
			{\includegraphics[width=0.26\textwidth,height=0.14\textheight]{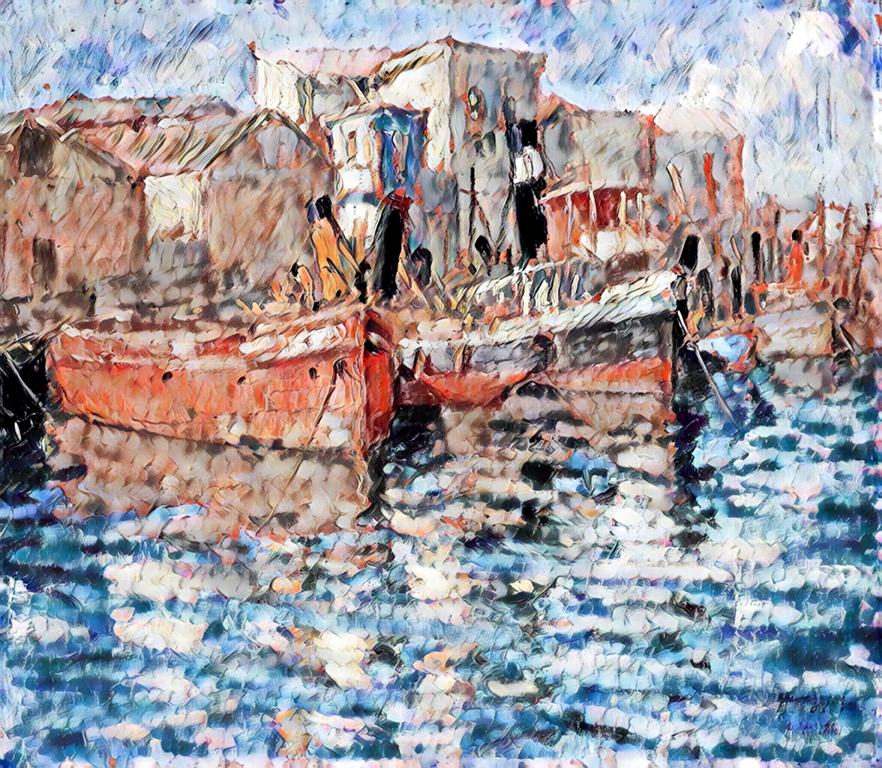}}\hfil
			{\includegraphics[width=0.2\textwidth,height=0.14\textheight]{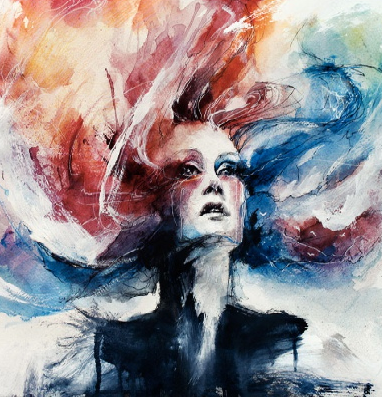}}\hfil
			\subfloat[ Input ]
			{\includegraphics[width=0.26\textwidth,height=0.14\textheight]{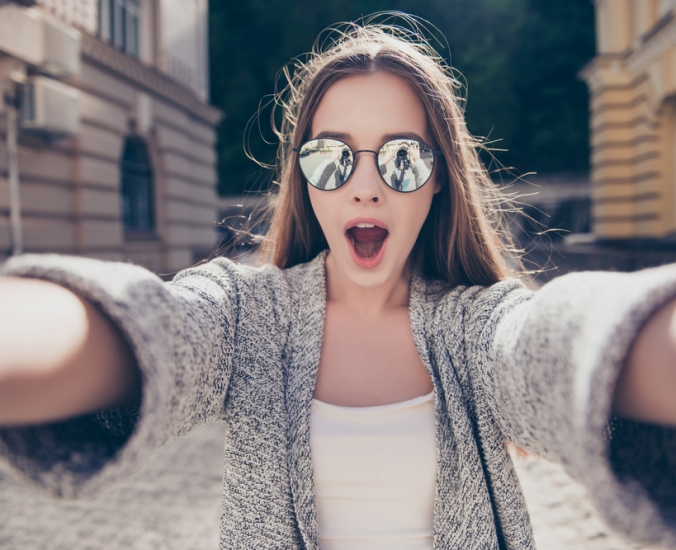}\label{fig:fig9a}}\hfil
			\subfloat[ IIT+ BF (Ours) ]
			{\includegraphics[width=0.26\textwidth,height=0.14\textheight]{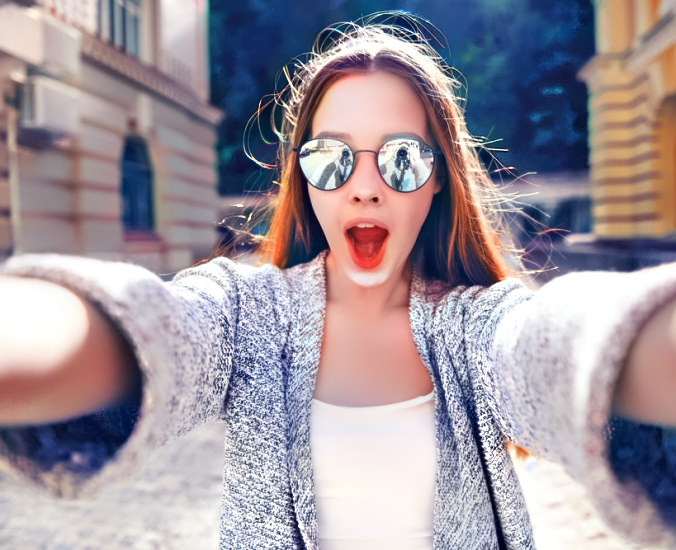}\label{fig:fig9b}}\hfil
			\subfloat[ Stylizied exemplar ]
			{\includegraphics[width=0.26\textwidth,height=0.14\textheight]{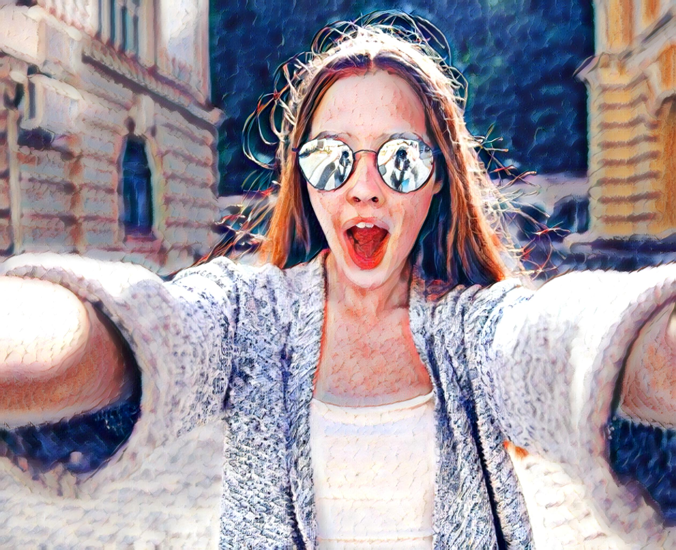}\label{fig:fig9c}}\hfil
			\subfloat[ Reference style]
			{\includegraphics[width=0.2\textwidth,height=0.14\textheight]{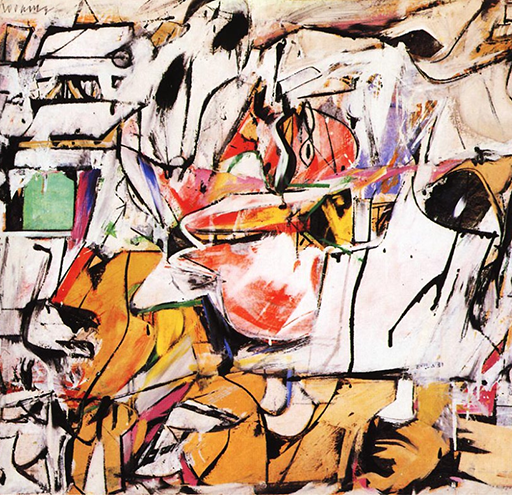}\label{fig:fig9d}}\hfil	
		\end{minipage}
	\end{center}
	\caption{Photorealistic style transfer. Given an image (a) and a reference style (d), a stylized exemplar (c) may be provided by these deep-learning methods\cite{gatys2016image, huang2017arbitrary, jing2019neural, li2017universal}, which is further refined by our IIT model as shown in (b) with more consistent textures and structures.}
	\label{fig:fig9}
\end{figure*}

In general, image quality assessment can be categorized into full-reference and no-reference approaches. The full-reference methods always require the corresponding ground-truth images for high-accuracy assessment, which, however, are always not available in the case of illumination-related tasks. As a result, it is difficult to take an objective full-reference evaluation. In contrast, no-reference methods require no ground-truth images but rely on statistical models to measure the degradation of an image. Recent no-reference approaches, in particular these deep-learning ones, have also shown promising success in predicting the quality of images. Specifically, we employ a quantitative evaluation based on the Tone Mapped Image Quality Index (TMQI)\cite{yeganeh2013objective}, Integrated Local Natural Image Quality Evaluator (IL-NIQE)\cite{zhang2015feature} and Neural Image Assessment (NIMA)\cite{talebi2018nima}. The TMQI\cite{yeganeh2013objective} index is a full-reference image assessment method built on the source and output images, which is originally proposed to provide an objective image quality assessment for HDR image compression. In TMQI index, Structural Fidelity (SF) and Statistical Naturalness (SN) are considered to provide an objective quality assessment. The SF index is based on the multi-scale structural similarity (SSIM) approach\cite{wang2004image} to extract structural information from the visual scene and provides a perceptual predictor of structural fidelity. The SN index provides a statistic metric for the output image, which takes the natural image statistics of brightness, contrast, visibility and details into account. IL-NIQE\cite{zhang2015feature} and  NIMA\cite{talebi2018nima} are two no-reference image assessments. The former is a learning method based on natural image statistics features derived from color, luminance,  gradient and structure information; and the latter attempts to predict consistent aesthetic scores with human opinions using convolutional neural networks. The NIMA\cite{talebi2018nima} method is trained on a large-scale natural image dataset for perceptually-aware no-reference quality assessment. The statistical results are shown in \textbf{Table} \ref{tab:tab1}, where the best two results are highlighted with \textbf{bold} and \textbf{underline}, respectively. The advantage of our IIT method is noticeable with 19 top-two places in total 30 indices compared with the state-of-art on real-world image datasets.

\subsection{More Extensions}

It is easy to see the proposed IIT method can be extended to high dynamic range (HDR) image compression to improve the visibility of dark regions. Similarly, it helps to suppress the distorted details or artifacts, providing the comparable or superior results than the prevailing cutting-edge methods. In this case, we apply the proposed IIT model on image luminance with the same aforementioned configurations. The saturation is restored with a heuristic de-saturation setup as described in\cite{fattal2002gradient}.
\begin{equation}
	\begin{aligned}
		C_{out}={\left(\frac{C_{in}} {L_{in}}\right)}^s L_{out},
	\end{aligned}
	\label{f2}
\end{equation}
where ${C =\{ R,G,B\}}$ are red, green and blue channels of color images, ${L_{in}}$ and ${L_{out}}$ denote the source and mapped luminance, and ${s}$ controls the saturation with an empirical value between 0.4 and 0.6 to produce satisfying results.

We show two examples of HDR image compression with visual comparisons in Fig.  \ref{fig:fig8}, where the exemplars are generated by the CLAHE algorithm\cite{Zuiderveld1994Contrast}. We set the parameter $s = 0.5$ and $s=0.6$ for the two cases, respectively. As we can see, the proposed IIT method is capable of producing high-quality results as that of the gradient-domain (GD) compression\cite{fattal2002gradient} and weighted least square (WLS) filtering method\cite{farbman2008edge}.  

We additionally illustrate that the structure-consistent property of the proposed IIT method makes it applicable to much more complex situations beyond the above illumination-related tasks. Specifically, it is possible to be used as a building block for color transform\cite{reinhard2001color} and style transfer\cite{gatys2016image, jing2019neural, li2017universal}, where a high-level vision image-to-image translation is always deduced to produce stylized results for artistic purpose. In the literature, many approaches, especially the deep-learning ones\cite{jing2019neural, li2017universal}, have been explored to obtain artistic-like stylized results. Despite their great success, they occasionally produce unexpected distortions or unrealistic artifacts that not occur in real photographs. We illustrate that the proposed IIT model is possible to suppress these non-consistent textures, enjoying a photo-realistic style transfer as proposed in the recent work\cite{lu2019closed, yoo2019photorealistic}. The understanding for style transfer, somewhat, remains elusive\cite{jing2019neural} and the discussion for more details is out of scope here, we conjecture here that the image-to-image translation procedure can be also interpreted under the intrinsic images of Eq. (1).  The difference from the illumination-related works is that both illumination and reflectance layers in style transfer vary significantly, leading to a strong discrepancy in image color, saturation, texture, style, and so on.

We briefly illustrate the procedure by giving an input image and a reference style, where a stylized exemplar is firstly generated by the existing style transfer methods such as the deep learning ones\cite{gatys2016image, huang2017arbitrary, jing2019neural, li2017universal}.  Our IIT algorithm is then used to refine the local distortions of the stylized exemplar. As shown in Fig.  \ref{fig:fig9}, an image Fig.  \ref{fig:fig9a} is first transformed into the stylized exemplar Fig.  \ref{fig:fig9c} using two deep learning-based methods\cite{gatys2016image, jing2019neural} under the guidance of reference style Fig.  \ref{fig:fig9d}; and a photo-realistic result Fig.  \ref{fig:fig9b} is obtained under the IIT model with the stylized image as the corresponding exemplar. As we can see,  our IIT algorithm produces satisfying results  that have identical saturation as the exemplar and high-quality consistent structures as the input image.  The local textures in the ``sky'' region are remarkably removed, leading to the photo-realistic stylized results. The startling results demonstrate the powerful ability of the proposed IIT method in suppressing the local distortions, and it further turns out the robustness of our method to the exemplar image.

\section{Conclusion}\label{conclusion}  
This paper has described an intrinsic image transfer algorithm,  which is rather different from recent trends towards making an intrinsic image decomposition.  This model creates a local image translation between two image surfaces and produces high-quality results in a wide range of illumination manipulation tasks.  One drawback is that the algorithm is time-consuming for the need of computing the large-scale LLE weights and PCG solver. It is possible to be addressed with a sub-sampling strategy for efficiency, which is left for future work.


%

\appendices


\ifCLASSOPTIONcompsoc
  \section*{Acknowledgments}
   This work was in part supported by the National Science and Technology Major Project, China (Nos. 2019-I-0001-0001 and 2019-I-0019-0018) and Shandong MSTI Project, China (No. 2019JZZY010122), and Foundation for Innovative Young Talents in Higher Education of Guangdong, China (No. 2021KQNCX213). This work was also partially supported by the FWO Odysseus Project, Leverhulme Grant RPG-2017-151 and EPSRC grant EP/R003025/1. We thank the anonymous reviewers for their careful reading of our manuscript and their insightful comments and suggestions. We also appreciate the related online resources, including images, codes, software, and so on. 
\else
  \section*{Acknowledgment}
\fi

\ifCLASSOPTIONcaptionsoff
  \newpage
\fi




\begin{thebibliography}{99}

\bibitem{land1971lightness}
E.~H. Land and J.~J. McCann, ``Lightness and retinex theory,'' \emph{Josa},
vol.~61, no.~1, pp. 1--11, 1971.

\bibitem{gilchrist2006seeing}
A.~Gilchrist, \emph{Seeing black and white}.\hskip 1em plus 0.5em minus
0.4em\relax OUP USA, 2006.

\bibitem{barrow1978recovering}
H.~Barrow and J.~Tenenbaum, ``Recovering intrinsic scene characteristics,''
\emph{Comput. Vis. Syst}, vol.~2, 1978.

\bibitem{mantiuk2015high}
R.~K. Mantiuk, K.~Myszkowski, and H.-P. Seidel, \emph{High dynamic range
	imaging}.\hskip 1em plus 0.5em minus 0.4em\relax Wiley Online Library, 2015.

\bibitem{morovic2001fundamentals}
J.~Morovic and M.~R. Luo, ``The fundamentals of gamut mapping: A survey,''
\emph{Journal of Imaging Science and Technology}, vol.~45, no.~3, pp.
283--290, 2001.

\bibitem{fattal2002gradient}
R.~Fattal, D.~Lischinski, and M.~Werman, ``Gradient domain high dynamic range
compression,'' in \emph{ACM transactions on graphics (TOG)}, vol.~21.\hskip
1em plus 0.5em minus 0.4em\relax ACM, 2002, pp. 249--256.

\bibitem{ng2011total}
M.~K. Ng and W.~Wang, ``A total variation model for retinex,'' \emph{SIAM
	Journal on Imaging Sciences}, vol.~4, no.~1, pp. 345--365, 2011.

\bibitem{roweis2000nonlinear}
S.~T. Roweis and L.~K. Saul, ``Nonlinear dimensionality reduction by locally
linear embedding,'' \emph{Science}, vol. 290, no. 5500, pp. 2323--2326, 2000.

\bibitem{land1977retinex}
E.~H. Land, ``The retinex theory of color vision,'' \emph{Scientific American},
vol. 237, no.~6, pp. 108--129, 1977.

\bibitem{horn1974determining}
B.~K. Horn, ``Determining lightness from an image,'' \emph{Computer graphics
	and image processing}, vol.~3, no.~4, pp. 277--299, 1974.

\bibitem{land1983recent}
E.~H. Land, ``Recent advances in retinex theory and some implications for
cortical computations: color vision and the natural image.''
\emph{Proceedings of the National Academy of Sciences of the United States of
	America}, vol.~80, no.~16, p. 5163, 1983.

\bibitem{jobson1997multiscale}
D.~J. Jobson, Z.-u. Rahman, and G.~A. Woodell, ``A multiscale retinex for
bridging the gap between color images and the human observation of scenes,''
\emph{IEEE Transactions on Image processing}, vol.~6, no.~7, pp. 965--976,
1997.

\bibitem{kimmel2003variational}
R.~Kimmel, M.~Elad, D.~Shaked, R.~Keshet, and I.~Sobel, ``A variational
framework for retinex,'' \emph{International Journal of computer
	vision(IJCV)}, vol.~52, no.~1, pp. 7--23, 2003.

\bibitem{morel2010pde}
J.~M. Morel, A.~B. Petro, and C.~Sbert, ``A pde formalization of retinex
theory,'' \emph{IEEE Transactions on Image Processing}, vol.~19, no.~11, pp.
2825--2837, 2010.

\bibitem{elad2005retinex}
M.~Elad, ``Retinex by two bilateral filters,'' in \emph{International
	Conference on Scale-Space Theories in Computer Vision}.\hskip 1em plus 0.5em
minus 0.4em\relax Springer, 2005, pp. 217--229.

\bibitem{tomasi1998bilateral}
C.~Tomasi and R.~Manduchi, ``Bilateral filtering for gray and color images,''
in \emph{Computer Vision, 1998. Sixth International Conference on}.\hskip 1em
plus 0.5em minus 0.4em\relax IEEE, 1998, pp. 839--846.

\bibitem{farbman2008edge}
Z.~Farbman, R.~Fattal, D.~Lischinski, and R.~Szeliski, ``Edge-preserving
decompositions for multi-scale tone and detail manipulation,'' in \emph{ACM
	Transactions on Graphics (TOG)}, vol.~27.\hskip 1em plus 0.5em minus
0.4em\relax ACM, 2008, p.~67.

\bibitem{milanfar2012tour}
P.~Milanfar, ``A tour of modern image filtering: New insights and methods, both
practical and theoretical,'' \emph{IEEE signal processing magazine}, vol.~30,
no.~1, pp. 106--128, 2012.

\bibitem{rother2011recovering}
C.~Rother, M.~Kiefel, L.~Zhang, B.~Sch{\"o}lkopf, and P.~V. Gehler,
``Recovering intrinsic images with a global sparsity prior on reflectance,''
in \emph{Proceedings of Advances in neural information processing systems
	(NIPS)}, 2011, pp. 765--773.

\bibitem{bell2014intrinsic}
S.~Bell, K.~Bala, and N.~Snavely, ``Intrinsic images in the wild,'' \emph{ACM
	Transactions on Graphics (TOG)}, vol.~33, no.~4, p. 159, 2014.

\bibitem{lombardi2016radiometric}
S.~Lombardi and K.~Nishino, ``Radiometric scene decomposition: Scene
reflectance, illumination, and geometry from rgb-d images,'' in \emph{3D
	Vision (3DV), 2016 Fourth International Conference on}.\hskip 1em plus 0.5em
minus 0.4em\relax IEEE, 2016, pp. 305--313.

\bibitem{jeon2014intrinsic}
J.~Jeon, S.~Cho, X.~Tong, and S.~Lee, ``Intrinsic image decomposition using
structure-texture separation and surface normals,'' in \emph{In Proceedings
	of the European Conference on Computer Vision (ECCV)}.\hskip 1em plus 0.5em
minus 0.4em\relax Springer, 2014, pp. 218--233.

\bibitem{barron2015shape}
J.~T. Barron and J.~Malik, ``Shape, illumination, and reflectance from
shading,'' \emph{IEEE transactions on pattern analysis and machine
	intelligence}, vol.~37, no.~8, pp. 1670--1687, 2015.

\bibitem{guo2017lime}
X.~Guo, Y.~Li, and H.~Ling, ``Lime: Low-light image enhancement via
illumination map estimation,'' \emph{IEEE transactions on image processing},
vol.~26, no.~2, pp. 982--993, 2017.

\bibitem{Cordts2016Cityscapes}
M.~Cordts, M.~Omran, S.~Ramos, T.~Rehfeld, M.~Enzweiler, R.~Benenson,
U.~Franke, S.~Roth, and B.~Schiele, ``The cityscapes dataset for semantic
urban scene understanding,'' in \emph{Proceedings of the IEEE Conference on
	Computer Vision and Pattern Recognition (CVPR)}, 2016.

\bibitem{geiger2013vision}
A.~Geiger, P.~Lenz, C.~Stiller, and R.~Urtasun, ``Vision meets robotics: The
kitti dataset,'' \emph{The International Journal of Robotics Research},
vol.~32, no.~11, pp. 1231--1237, 2013.

\bibitem{ignatov2017dslr}
A.~Ignatov, N.~Kobyshev, K.~Vanhoey, R.~Timofte, and L.~Van~Gool,
``Dslr-quality photos on mobile devices with deep convolutional networks,''
in \emph{Proceedings of the International Conference on Computer Vision
	(ICCV)}, 2017.

\bibitem{krizhevsky2012imagenet}
A.~Krizhevsky, I.~Sutskever, and G.~E. Hinton, ``Imagenet classification with
deep convolutional neural networks,'' in \emph{Proceedings of Advances in
	neural information processing systems (NIPS)}, 2012, pp. 1097--1105.

\bibitem{goodfellow2014generative}
I.~Goodfellow, J.~Pouget-Abadie, M.~Mirza, B.~Xu, D.~Warde-Farley, S.~Ozair,
A.~Courville, and Y.~Bengio, ``Generative adversarial nets,'' in
\emph{Proceedings of Advances in neural information processing systems
	(NIPS)}, 2014, pp. 2672--2680.

\bibitem{cheng2018intrinsic}
L.~Cheng, C.~Zhang, and Z.~Liao, ``Intrinsic image transformation via scale
space decomposition,'' in \emph{Proceedings of the IEEE Conference on
	Computer Vision and Pattern Recognition (CVPR)}, 2018, pp. 656--665.

\bibitem{gardner2017learning}
M.-A. Gardner, K.~Sunkavalli, E.~Yumer, X.~Shen, E.~Gambaretto, C.~Gagn{\'e},
and J.-F. Lalonde, ``Learning to predict indoor illumination from a single
image,'' \emph{ACM Transactions on Graphics (TOG)}, vol.~36, no.~6, p. 176,
2017.

\bibitem{hold2017deep}
Y.~Hold-Geoffroy, K.~Sunkavalli, S.~Hadap, E.~Gambaretto, and J.-F. Lalonde,
``Deep outdoor illumination estimation,'' in \emph{Proceedings of the IEEE
	Conference on Computer Vision and Pattern Recognition (CVPR)}, vol.~2, 2017.

\bibitem{li2018learning}
Z.~Li and N.~Snavely, ``Learning intrinsic image decomposition from watching
the world,'' in \emph{Proceedings of the IEEE Conference on Computer Vision
	and Pattern Recognition (CVPR)}, 2018, pp. 9039--9048.

\bibitem{meka2018lime}
A.~Meka, M.~Maximov, M.~Zollh{\"o}fer, A.~Chatterjee, H.-P. Seidel,
C.~Richardt, and C.~Theobalt, ``Lime: Live intrinsic material estimation,''
in \emph{Proceedings of the IEEE Conference on Computer Vision and Pattern
	Recognition (CVPR)}, 2018, pp. 6315--6324.

\bibitem{wei2018deep}
C.~Wei, W.~Wang, W.~Yang, and J.~Liu, ``Deep retinex decomposition for
low-light enhancement,'' \emph{arXiv preprint arXiv:1808.04560}, 2018.

\bibitem{narihira2015direct}
T.~Narihira, M.~Maire, and S.~X. Yu, ``Direct intrinsics: Learning
albedo-shading decomposition by convolutional regression,'' in
\emph{Proceedings of the International Conference on Computer Vision (ICCV)},
2015, pp. 2992--2992.

\bibitem{butler2012naturalistic}
D.~J. Butler, J.~Wulff, G.~B. Stanley, and M.~J. Black, ``A naturalistic open
source movie for optical flow evaluation,'' in \emph{European conference on
	computer vision}.\hskip 1em plus 0.5em minus 0.4em\relax Springer, 2012, pp.
611--625.

\bibitem{chen2018learning}
C.~Chen, Q.~Chen, J.~Xu, and V.~Koltun, ``Learning to see in the dark,''
\emph{arXiv preprint arXiv:1805.01934}, 2018.

\bibitem{ignatov2017wespe}
A.~Ignatov, N.~Kobyshev, R.~Timofte, K.~Vanhoey, and L.~Van~Gool, ``Wespe:
weakly supervised photo enhancer for digital cameras,'' in \emph{Proceedings
	of the IEEE Conference on Computer Vision and Pattern Recognition Workshops},
2018, pp. 691--700.

\bibitem{guo2020zero}
C.~Guo, C.~Li, J.~Guo, C.~C. Loy, J.~Hou, S.~Kwong, and R.~Cong,
``Zero-reference deep curve estimation for low-light image enhancement,'' in
\emph{Proceedings of the IEEE/CVF Conference on Computer Vision and Pattern
	Recognition}, 2020, pp. 1780--1789.

\bibitem{chen2018deep}
Y.-S. Chen, Y.-C. Wang, M.-H. Kao, and Y.-Y. Chuang, ``Deep photo enhancer:
Unpaired learning for image enhancement from photographs with gans,'' in
\emph{Proceedings of the IEEE Conference on Computer Vision and Pattern
	Recognition}, 2018, pp. 6306--6314.

\bibitem{janner2017self}
M.~Janner, J.~Wu, T.~D. Kulkarni, I.~Yildirim, and J.~Tenenbaum,
``Self-supervised intrinsic image decomposition,'' \emph{Advances in Neural
	Information Processing Systems}, vol.~30, 2017.

\bibitem{jiang2019enlightengan}
Y.~Jiang, X.~Gong, D.~Liu, Y.~Cheng, C.~Fang, X.~Shen, J.~Yang, P.~Zhou, and
Z.~Wang, ``Enlightengan: Deep light enhancement without paired supervision,''
\emph{arXiv preprint arXiv:1906.06972}, 2019.

\bibitem{zhu2017unpaired}
J.-Y. Zhu, T.~Park, P.~Isola, and A.~A. Efros, ``Unpaired image-to-image
translation using cycle-consistent adversarial networks,'' in
\emph{Proceedings of the IEEE international conference on computer vision},
2017, pp. 2223--2232.

\bibitem{zhang2021star}
Z.~Zhang, Y.~Jiang, J.~Jiang, X.~Wang, P.~Luo, and J.~Gu, ``Star: A
structure-aware lightweight transformer for real-time image enhancement,'' in
\emph{Proceedings of the IEEE/CVF International Conference on Computer
	Vision}, 2021, pp. 4106--4115.

\bibitem{shen2011intrinsic}
L.~Shen and C.~Yeo, ``Intrinsic images decomposition using a local and global
sparse representation of reflectance,'' in \emph{CVPR 2011}.\hskip 1em plus
0.5em minus 0.4em\relax IEEE, 2011, pp. 697--704.

\bibitem{huang2018multispectral}
Q.~Huang, W.~Zhu, Y.~Zhao, L.~Chen, Y.~Wang, T.~Yue, and X.~Cao,
``Multispectral image intrinsic decomposition via low rank constraint,''
\emph{arXiv preprint arXiv:1802.08793}, 2018.

\bibitem{zheng2015illumination}
Y.~Zheng, I.~Sato, and Y.~Sato, ``Illumination and reflectance spectra
separation of a hyperspectral image meets low-rank matrix factorization.'' in
\emph{Proceedings of the IEEE Conference on Computer Vision and Pattern
	Recognition (CVPR)}, 2015, pp. 1779--1787.

\bibitem{zhao2012closed}
Q.~Zhao, P.~Tan, Q.~Dai, L.~Shen, E.~Wu, and S.~Lin, ``A closed-form solution
to retinex with nonlocal texture constraints,'' \emph{IEEE transactions on
	pattern analysis and machine intelligence}, vol.~34, no.~7, pp. 1437--1444,
2012.

\bibitem{zosso2015non}
D.~Zosso, G.~Tran, and S.~J. Osher, ``Non-local retinex---a unifying framework
and beyond,'' \emph{SIAM Journal on Imaging Sciences}, vol.~8, no.~2, pp.
787--826, 2015.

\bibitem{Zuiderveld1994Contrast}
K.~Zuiderveld, ``Contrast limited adaptive histogram equalization,''
\emph{Graphics gems}, pp. 474--485, 1994.

\bibitem{saul2003think}
L.~K. Saul and S.~T. Roweis, ``Think globally, fit locally: unsupervised
learning of low dimensional manifolds,'' \emph{Journal of machine learning
	research}, vol.~4, no. Jun, pp. 119--155, 2003.

\bibitem{zhang2007mlle}
Z.~Zhang and J.~Wang, ``Mlle: Modified locally linear embedding using multiple
weights,'' in \emph{Proceedings of Advances in neural information processing
	systems (NIPS)}, 2007, pp. 1593--1600.

\bibitem{saad2003iterative}
Y.~Saad, \emph{Iterative methods for sparse linear systems}.\hskip 1em plus
0.5em minus 0.4em\relax SIAM, 2003, vol.~82.

\bibitem{schroff2011pose}
F.~Schroff, T.~Treibitz, D.~Kriegman, and S.~Belongie, ``Pose, illumination and
expression invariant pairwise face-similarity measure via doppelg{\"a}nger
list comparison,'' in \emph{Proceedings of the International Conference on
	Computer Vision (ICCV)}.\hskip 1em plus 0.5em minus 0.4em\relax IEEE, 2011,
pp. 2494--2501.

\bibitem{photoshop_cc}
``Adobe photoshop cc,''
\url{https://www.adobe.com/cn/products/photoshop.html/}.

\bibitem{nasa_retinex}
``Nasa retinex,'' \url{https://dragon.larc.nasa.gov/retinex/}.

\bibitem{google_nik}
``Google nikcollection,'' \url{https://www.google.com/nikcollection/}.

\bibitem{openexr_ward}
``High dynamic range image examples,''
\url{http://www.anyhere.com/gward/hdrenc/pages/originals.html}.

\bibitem{gatys2016image}
L.~A. Gatys, A.~S. Ecker, and M.~Bethge, ``Image style transfer using
convolutional neural networks,'' in \emph{Proceedings of the IEEE conference
	on computer vision and pattern recognition}, 2016, pp. 2414--2423.

\bibitem{huang2017arbitrary}
X.~Huang and S.~Belongie, ``Arbitrary style transfer in real-time with adaptive
instance normalization,'' in \emph{Proceedings of the IEEE international
	conference on computer vision}, 2017, pp. 1501--1510.

\bibitem{jing2019neural}
Y.~Jing, Y.~Yang, Z.~Feng, J.~Ye, Y.~Yu, and M.~Song, ``Neural style transfer:
A review,'' \emph{IEEE transactions on visualization and computer graphics},
vol.~26, no.~11, pp. 3365--3385, 2019.

\bibitem{li2017universal}
Y.~Li, C.~Fang, J.~Yang, Z.~Wang, X.~Lu, and M.-H. Yang, ``Universal style
transfer via feature transforms,'' \emph{Advances in neural information
	processing systems}, vol.~30, 2017.

\bibitem{yeganeh2013objective}
H.~Yeganeh and Z.~Wang, ``Objective quality assessment of tone-mapped images,''
\emph{IEEE transactions on image processing}, vol.~22, no.~2, pp. 657--667,
2013.

\bibitem{zhang2015feature}
L.~Zhang, L.~Zhang, and A.~C. Bovik, ``A feature-enriched completely blind
image quality evaluator,'' \emph{IEEE transactions on image processing},
vol.~24, no.~8, pp. 2579--2591, 2015.

\bibitem{talebi2018nima}
H.~Talebi and P.~Milanfar, ``Nima: Neural image assessment,'' \emph{IEEE
	Transactions on Image Processing}, vol.~27, no.~8, pp. 3998--4011, 2018.

\bibitem{wang2004image}
Z.~Wang, A.~C. Bovik, H.~R. Sheikh, and E.~P. Simoncelli, ``Image quality
assessment: from error visibility to structural similarity,'' \emph{IEEE
	transactions on image processing}, vol.~13, no.~4, pp. 600--612, 2004.

\bibitem{reinhard2001color}
E.~Reinhard, M.~Adhikhmin, B.~Gooch, and P.~Shirley, ``Color transfer between
images,'' \emph{IEEE Computer graphics and applications}, vol.~21, no.~5, pp.
34--41, 2001.

\bibitem{lu2019closed}
M.~Lu, H.~Zhao, A.~Yao, Y.~Chen, F.~Xu, and L.~Zhang, ``A closed-form solution
to universal style transfer,'' in \emph{Proceedings of the IEEE/CVF
	International Conference on Computer Vision}, 2019, pp. 5952--5961.

\bibitem{yoo2019photorealistic}
J.~Yoo, Y.~Uh, S.~Chun, B.~Kang, and J.-W. Ha, ``Photorealistic style transfer
via wavelet transforms,'' in \emph{Proceedings of the IEEE/CVF International
	Conference on Computer Vision}, 2019, pp. 9036--9045.

\end{thebibliography}
%

%

\begin{IEEEbiography}[{\includegraphics[width=1in,height=1.25in,clip,keepaspectratio]{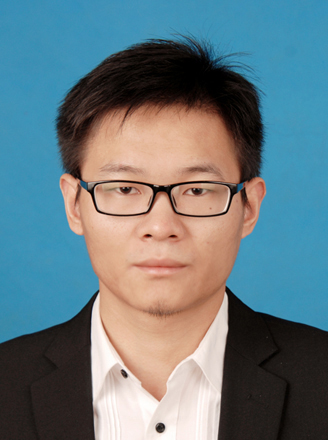}}]{Junqing Huang}
	received the BS degree in Automation from the School of Electrical Engineering, Zhengzhou University, Zhengzhou, China, in 2011, and the MS degree in Mathematics from the School of Mathematical Sciences, Beihang University (BUAA), Beijing, China, in 2015. He is currently a Ph.D. candidate of Department of Mathematics: Analysis, Logic and Discrete Mathematics, Ghent University, Belgium. His research interests include deep learning, image processing, optimal transport and optimization.
\end{IEEEbiography}

\vspace{-6mm}
\begin{IEEEbiography}[{\includegraphics[width=1in,height=1.25in,clip,keepaspectratio]{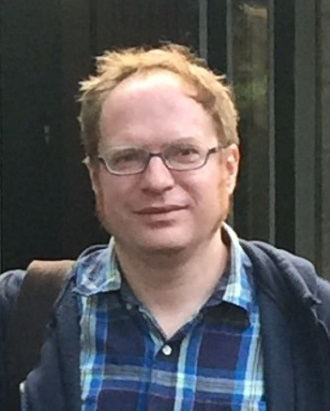}}]{Michael Ruzhansky}
	is a senior full Professor in Department of Mathematics, Ghent University, Belgium, a Professorship in School of Mathematical Sciences, Queen Mary University of London, and Honorary Professorship in Department of Mathematics, Imperial College London, UK.	He was awarded by FWO (Belgium) the prestigious Odysseus 1 Project in 2018, he was recipient of several Prizes and Awards: ISAAC Award in 2007,
	Daiwa Adrian Prize in 2010 and Ferran Sunyer I Balaguer Prizes in 2014 and 2018. His research interests include different areas of analysis, in particular, theory of PDEs, microlocal analysis, and harmonic analysis.
\end{IEEEbiography}

\vspace{-6mm}
\begin{IEEEbiography}[{\includegraphics[width=1in,height=1.25in,clip,keepaspectratio]{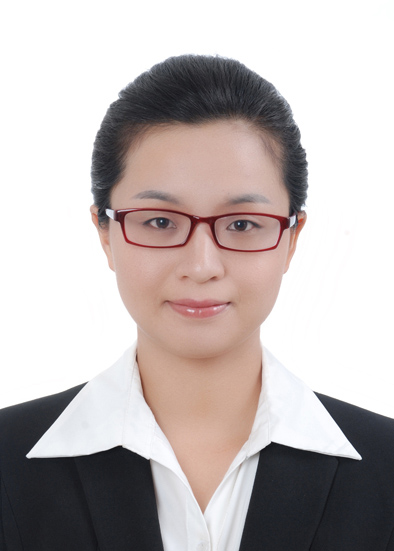}}]{Qianying Zhang}
	received the BS degree in Mathematics from Zhengzhou University, Zhengzhou, China, in 2011, and the PhD degree in Mathematics from Beihang University (BUAA), Beijing, China, in 2015. She is in Shenzhen Institute of Information Technology. Her research interests include signal and image processing, compressed sensing, sparse representation and artificial intelligence.
\end{IEEEbiography}

\vspace{-6mm}
\begin{IEEEbiography}[{\includegraphics[width=1in,height=1.25in,clip,keepaspectratio]{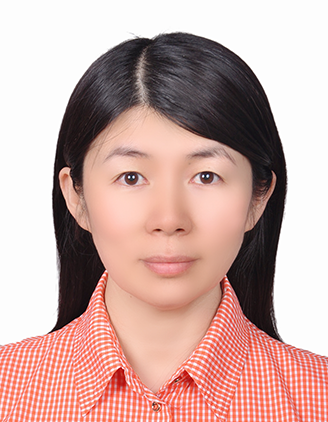}}]{Haihui Wang}	
	received the PhD degree in Mathematics from Peking University, Beijing, China, in 2003. She is currently a full Professor in the School of Mathematical Sciences, Beihang University (BUAA), Beijing, China, and a Professor in the Health Science Center, Beijing University, Beijing, China. Her research interests include artificial intelligence, machine learning, signal and image processing, wavelet analysis and application, and so on.
\end{IEEEbiography}
\vfill

\end{document}